\documentclass[letterpaper, 10 pt, journal, twoside]{IEEEtran} 

\IEEEoverridecommandlockouts                              

\usepackage[dvipsnames]{xcolor}
\definecolor{dark-green}{RGB}{12,80,12}

\usepackage{times}
\usepackage{multicol}
\usepackage[breaklinks,colorlinks]{hyperref}
\usepackage{cuted} 
\usepackage[dvipsnames]{xcolor}
\usepackage{graphicx}
\usepackage{amsmath}
\usepackage{amssymb}
\usepackage{booktabs}
\usepackage{microtype}
\usepackage[font={footnotesize}]{caption}
\usepackage{enumitem}
\usepackage{array}
\usepackage{flushend}
\usepackage{subcaption}
\usepackage{graphicx}
\usepackage{multirow}
\usepackage{mathtools}
\usepackage{xcolor}
\usepackage[export]{adjustbox}
\usepackage[hang,flushmargin]{footmisc}
\usepackage[resetlabels]{multibib}
\newcites{New}{References}

\newcommand{\secref}[1]{Sec.~\ref{#1}}
\newcommand{\figref}[1]{Fig.~\ref{#1}}
\newcommand{\tabref}[1]{Tab.~\ref{#1}}
 
\newcommand{\net}{PoDS}

\newcolumntype{P}[1]{>{\centering\arraybackslash}p{#1}}
\newcommand{\set}[1]{\left\{#1\right\}}
\newcommand{\nset}{\set}
\newcommand{\rot}[1]{\rotatebox[origin=c]{90}{#1}}

\renewcommand{\baselinestretch}{0.99}

\begin{document}

\title{\LARGE \bf
Panoptic Out-of-Distribution Segmentation}
\author{Rohit Mohan, 
        Kiran Kumaraswamy, 
        Juana Valeria Hurtado, 	
        Kürsat Petek, 
        and~Abhinav Valada%
        \thanks{Department of Computer Science, University of Freiburg, Germany.}%
        \thanks{This work was funded by the German Research Foundation (DFG) Emmy Noether Program grant No 468878300.}
}

\maketitle
\thispagestyle{empty}
\pagestyle{empty}

\begin{abstract}
Deep learning has led to remarkable strides in scene understanding with panoptic segmentation emerging as a key holistic scene interpretation task. However, the performance of panoptic segmentation is severely impacted in the presence of out-of-distribution (OOD) objects i.e. categories of objects that deviate from the training distribution. To overcome this limitation, we propose panoptic out-of-distribution segmentation for joint pixel-level semantic in-distribution and out-of-distribution classification with instance prediction. We extend two established panoptic segmentation benchmarks, Cityscapes and BDD100K, with out-of-distribution instance segmentation annotations, propose suitable evaluation metrics, and present multiple strong baselines. Importantly, we propose the novel PoDS architecture with a shared backbone, an OOD contextual module for learning global and local OOD object cues, and dual symmetrical decoders with task-specific heads that employ our alignment-mismatch strategy for better OOD generalization. Combined with our data augmentation strategy, this approach facilitates progressive learning of out-of-distribution objects while maintaining in-distribution performance. We perform extensive evaluations that demonstrate that our proposed PoDS network effectively addresses the main challenges and substantially outperforms the baselines.
We make the dataset, code, and trained models publicly available at \url{http://pods.cs.uni-freiburg.de}.
\end{abstract}


\section{Introduction}\label{sec:intro}
Recent advances in deep learning have substantially improved the capabilities of autonomous systems to interpret their surroundings~\cite{vodisch2022continual,gosala2023skyeye}. Central to these advancements is panoptic segmentation~\cite{kirillov2019panoptic}, which integrates semantic segmentation with instance segmentation, providing a holistic understanding of the environment. However, a significant challenge is that these models yield overconfident predictions of object categories out of the distribution they were trained on, known as out-of-distribution (OOD) objects. Segmenting these OOD objects poses a major challenge as they can vary significantly in appearance and semantics, include fine-grained details, and share visual characteristics with in-distribution objects, leading to ambiguity. Moreover, learning to jointly segment both OOD objects and in-distribution categories is extremely challenging as detailed in \secref{subsubsec:challenges}. Given the potential consequences of autonomous systems malfunctioning due to unexpected inputs~\cite{bozhinoski2019safety}, it is crucial to ensure the safe and robust deployment. 

\begin{figure}[t]
  \centering
  \includegraphics[width=0.7\linewidth,trim={0.0cm 0.0cm 0.0cm 0.0cm},clip,angle =0]{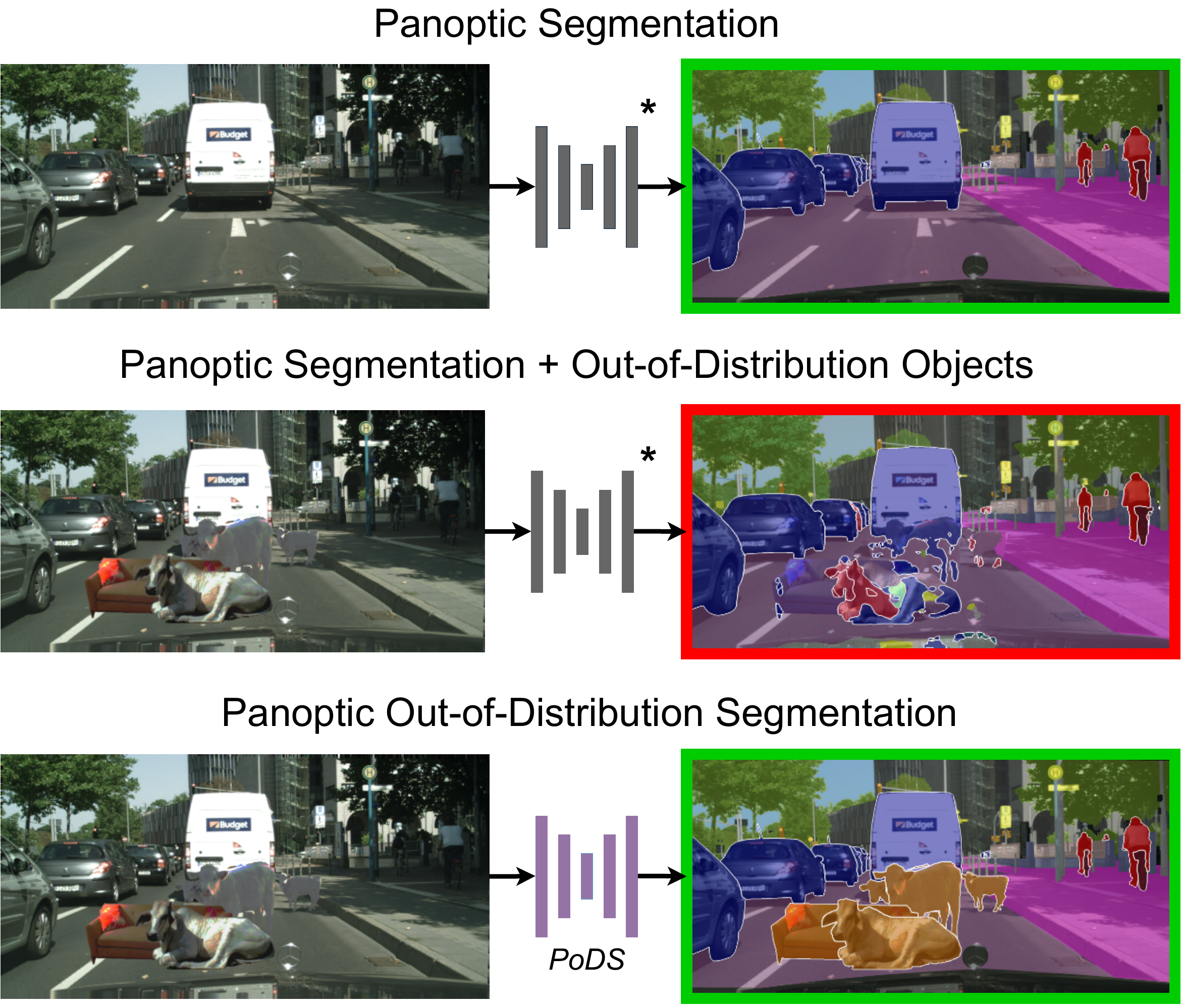}
   \caption{The \textit{panoptic segmentation} network (*) presents erroneous predictions when the input contains objects that are not representative of the distribution it was trained on. \textit{Panoptic out-of-distribution segmentation} aims to address this by predicting both semantic and instance segmentation of \textit{stuff} and \textit{thing} classes, while also predicting instances of unseen out-of-distribution classes.}
  \label{fig:teaser}
  \vspace{-0.4cm}
\end{figure}

To directly address these challenges at the task level, we introduce panoptic out-of-distribution segmentation that focuses on holistic scene understanding while effectively segmenting OOD objects. \figref{fig:teaser} illustrates our proposed task that aims to predict both the semantic segmentation of \textit{stuff} classes and instance segmentation of \textit{thing} classes as well as an OOD class. An object is considered OOD if it is not present in the training distribution but appears in the testing/deployment stages. This distinguishes panoptic OOD segmentation from the closely related open-set panoptic segmentation~\cite{hwang2021exemplar}. Further, panoptic OOD segmentation does not reason about the semantic differences between OOD objects since in most robotics settings, especially navigation, it is sufficient to identify OOD objects and further semantically categorizing them does not provide significant utility. 

In this work, we establish two challenging benchmarks, Cityscapes-OOD and BDD100K-OOD, by extending the standard autonomous driving datasets with OOD instance segmentation annotations. We present several strong baselines by combining semantic out-of-distribution segmentation methods with a class-agnostic instance segmentation decoder or adapting open-set segmentation approaches. We also introduce a tailored Panoptic Out-of-Distribution Quality (POD-Q) metric to quantify the performance. More importantly, as a first novel approach, we propose the PoDS architecture that incorporates out-of-distribution perception ability into a panoptic segmentation network conditioned on prior knowledge of in-distribution classes. By doing so, the network avoids the pitfalls of ambiguously modeling both OOD and in-distribution classes, thereby improving generalization and adept handling of unseen OOD objects. We perform extensive experimental evaluations that first demonstrate the feasibility of the task and further that our proposed PoDS architecture significantly outperforms the baselines, ensuring a balanced performance on both in-distribution and out-of-distribution classes.

In summary, the contributions of this work are as follows:
\begin{enumerate}
    \item We introduce the novel panoptic OOD segmentation task, identifying its main challenges, along with multiple baselines, and a suitable POD-Q metric. 
    \item We present the Cityscapes-OOD and BDD100K-OOD benchmarks, which extend the established datasets with OOD instance segmentation annotations.
    \item We propose the novel PoDS architecture that incorporates the proposed modules to embed OOD segmentation capabilities into a panoptic segmentation network leveraging conditional in-distribution priors.
    \item We present comprehensive quantitative and qualitative evaluations to demonstrate the feasibility of the task and the efficacy of our proposed PoDS architecture.
   \item We make the code, datasets, and models publicly available at \url{http://pods.cs.uni-freiburg.de}.
\end{enumerate}

\section{Related Work}\label{sec:relatedWork}
In this section, we present an overview of panoptic segmentation methods, followed by out-of-distribution segmentation approaches and open-set panoptic segmentation methods.


{\parskip=3pt
\noindent\textit{Panoptic Segmentation} methods can be categorized as top-down and bottom-up approaches. Top-down methods~\cite{mohan2020efficientps} employ task-specific heads, where the instance segmentation head predicts bounding boxes and corresponding masks for objects, while the semantic segmentation head generates dense semantic predictions for each class. The outputs from these heads are then combined using heuristic-based fusion modules~\cite{kirillov2019bpanoptic, mohan2020efficientps}. Conversely, bottom-up methods~\cite{cheng2020panoptic} begin with semantic segmentation and then employ various techniques~\cite{uhrig2018box2pix} to group \textit{thing} pixels together to obtain instance segmentation. Recently, Mohan~\textit{et~al.}~\cite{mohan22perceiving} introduced the PAPS architecture with a shared backbone, an asymmetrical dual-decoder, and several modules for amodal panoptic segmentation~\cite{mohan2022amodal}, which predicts both visible and occluded object segments. We base our approach on PAPS's modal variant, which perceives only visible segments as it outperforms other bottom-up methods.

}

{\parskip=3pt
\noindent\textit{Semantic Out-of-Distribution Segmentation} is often addressed through the use of uncertainty estimation techniques. A popular method is the maximum softmax probability (MSP)~\cite{hendrycks2016baseline} that uses probabilities from the softmax distribution. Following, maximum logit (MaxLogit)~\cite{hendrycks2019scaling} uses the negative
of the maximum unnormalized logit to deliver improved performance in semantic out-of-distribution segmentation over MSP. On the other hand, Bayesian networks generate uncertainty estimates by modeling their weights and outputs as probability distributions rather than fixed values~\cite{kendall2017uncertainties}. However, Bayesian inferences can be computationally expensive, hence in practice methods such as Dropout~\cite{gal2016dropout} or ensembles~\cite{lakshminarayanan2017simple} which capture model uncertainty by averaging predictions over multiple models are often used as approximations. Various frameworks also use density estimation~\cite{blum2019fishyscapes} via estimating the likelihood of samples with respect to the training distribution for addressing semantic out-of-distribution segmentation. Furthermore, \cite{chan2021entropy} proposes a loss function to yield high entropy for out-of-distribution sample predictions. The use of autoencoders on in-distribution data has also been explored to identify erroneous and less reliable reconstructions of out-of-distribution samples due to unseen patterns during training. Generative models~\cite{xia2020synthesize} generate OOD data as boundary samples. This is however very challenging to scale to complex and high-dimensional data such as high-resolution images of urban scenes. Other approaches include using adversarial perturbations on the input during training and test-time to predict in- and out-of-distribution samples. ODIN~\cite{liang2017enhancing} uses temperature scaling with small adversarial perturbations on the input at test-time, while~\cite{besnier2021triggering} use adversarial attacks during training as a proxy for out-of-distribution training samples. In this work, we adapt a number of aforementioned methods to serve as baselines for panoptic out-of-distribution segmentation as described in \secref{subsubsec:baselines}.

}

{\parskip=3pt
\noindent\textit{Open-Set Panoptic Segmentation}: There are only two approaches that have been proposed thus far. EOPSN~\cite{hwang2021exemplar} groups similar unlabeled objects across multiple inputs during training and assigns labels to unlabeled objects that are surrounded by known segments. Following, Xu~\textit{et~al.}~\cite{xu2022dual} use a known classification head to reject segments while employing a class-agnostic classifier to identify the segments as unknown objects. We use these methods as baselines with some adaptations as described in \secref{subsubsec:baselines}.

\section{Panoptic Out-of-Distribution Segmentation}

\subsubsection{Task Definition}
\label{subsubsec:taskdef}
Panoptic out-of-distribution segmentation aims to assign each pixel $i$ of an input image to an output pair $(c_i, \kappa_i) \in (C \cup O) \times N$. Here, $C$ denotes known semantic classes, while $O$ represents the out-of-distribution class, such that $C \cap O = \emptyset$, and $N$ is the total number of instances. $C$ is further divided into \textit{stuff} labels $C^S$ (e.g., sidewalks) and \textit{thing} labels $C^T$ (e.g., pedestrians). In this task, the variable $c_i$ can be a semantic or OOD class, and 
$\kappa_i$ indicates the corresponding instance ID. For \textit{stuff} classes, $\kappa_i$ is not applicable.

\subsubsection{Evaluation Metric}
\label{subsubsec:metric}
To quantify the performance, it is essential to evaluate both in-distribution and out-of-distribution performance equally. To this end, we introduce the Panoptic Out-of-Distribution Quality (POD-Q), which builds upon the panoptic quality (PQ) metric~\cite{kirillov2019panoptic}. We first determine $PQ_{out}$ representing the PQ for the $OOD$ class and $PQ_{in}$ accounting for all the in-distribution semantic classes. Finally, POD-Q is computed as the geometric mean of $PQ_{out}$ and $PQ_{in}$: 
\begin{equation}
POD\text{-}Q = (PQ_{out} \times PQ_{in})^{\frac{1}{2}}.
\end{equation}

We use the geometric mean to incentivize balanced performance in both out-of-distribution and in-distribution segmentation while strictly penalizing methods that only excel in one aspect of the task. For further details on $PQ_{out}$ and $PQ_{in}$, please refer to \secref{sec:smetric} of the supplementary material.

\subsubsection{Challenges}
\label{subsubsec:challenges}


Classifying and segmenting objects that do not belong to the known training distribution is challenging due to the absence of explicit knowledge about diverse OOD object characteristics. This becomes more challenging with the simultaneous identification and segmentation of OOD objects with panoptic segmentation of in-distribution classes. The increased complexity makes naive adaption of methods from the less complex tasks such as semantic out-of-distribution segmentation vulnerable to trade-offs prioritizing one aspect over the other. Unsupervised methods that condition segmentation outputs based on threshold scores to predict OOD objects become sensitive, as any fragmentation in predictions can result in false instance predictions. Conversely, in learning with supervised OOD data, where training data is limited and does not encompass all OOD object variations, models can overfit to specific OOD objects encountered during training. Consequently, this results in difficulties recognizing new OOD objects during inference. Considering the panoptic out-of-distribution task, which demands a balanced performance for both in-distribution and out-of-distribution scene elements, it becomes evident that a comprehensive approach is required. 

\begin{figure}
    \centering
    \begin{subfigure}[b]{0.49\linewidth}
        \centering
        \includegraphics[width=\textwidth]{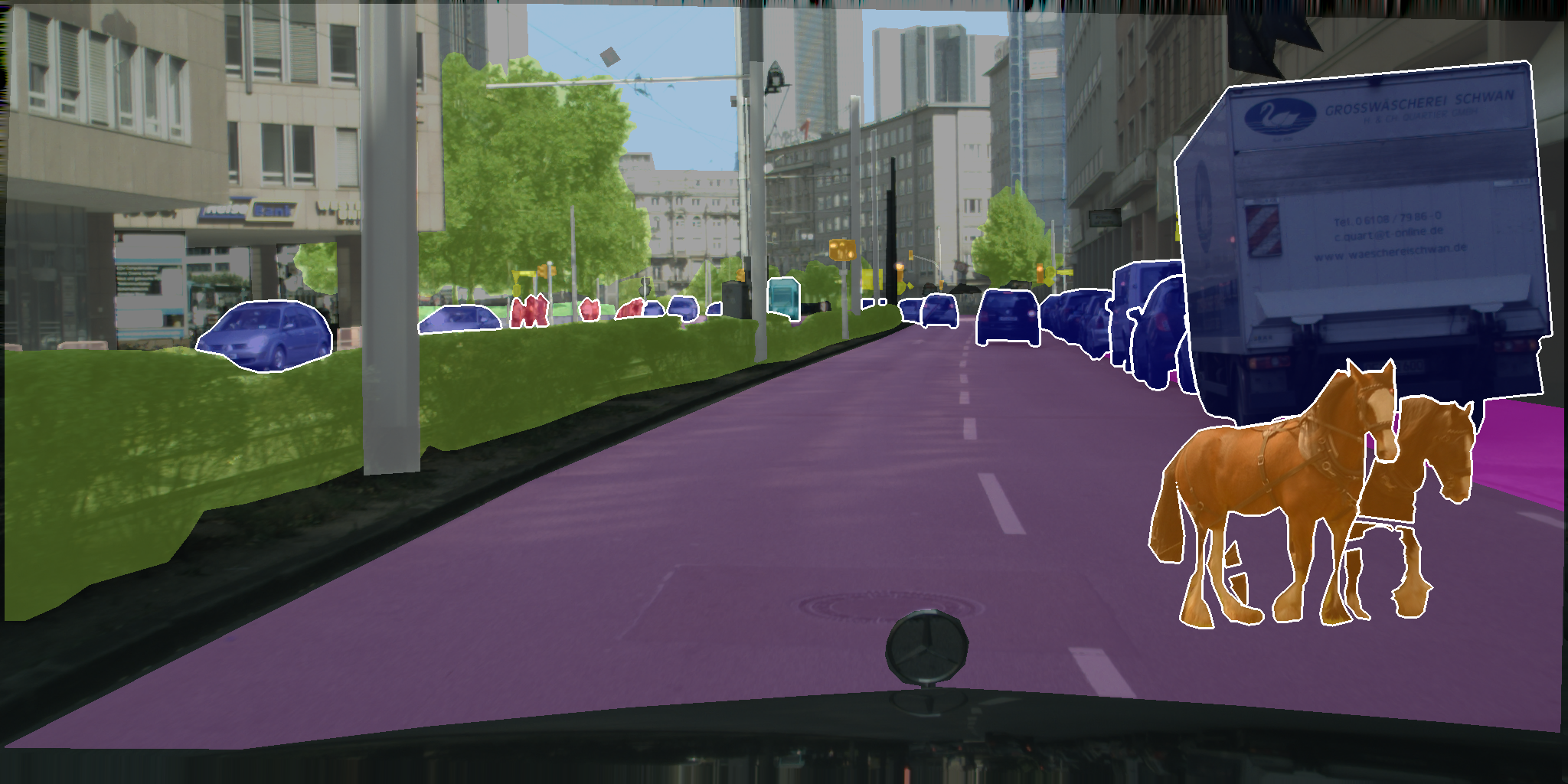} 
        \subcaption{Random scaling }
        \label{panoptic_segmentation_eg}
    \end{subfigure}
    \begin{subfigure}[b]{0.49\linewidth}
        \centering 
        \includegraphics[width=\textwidth]{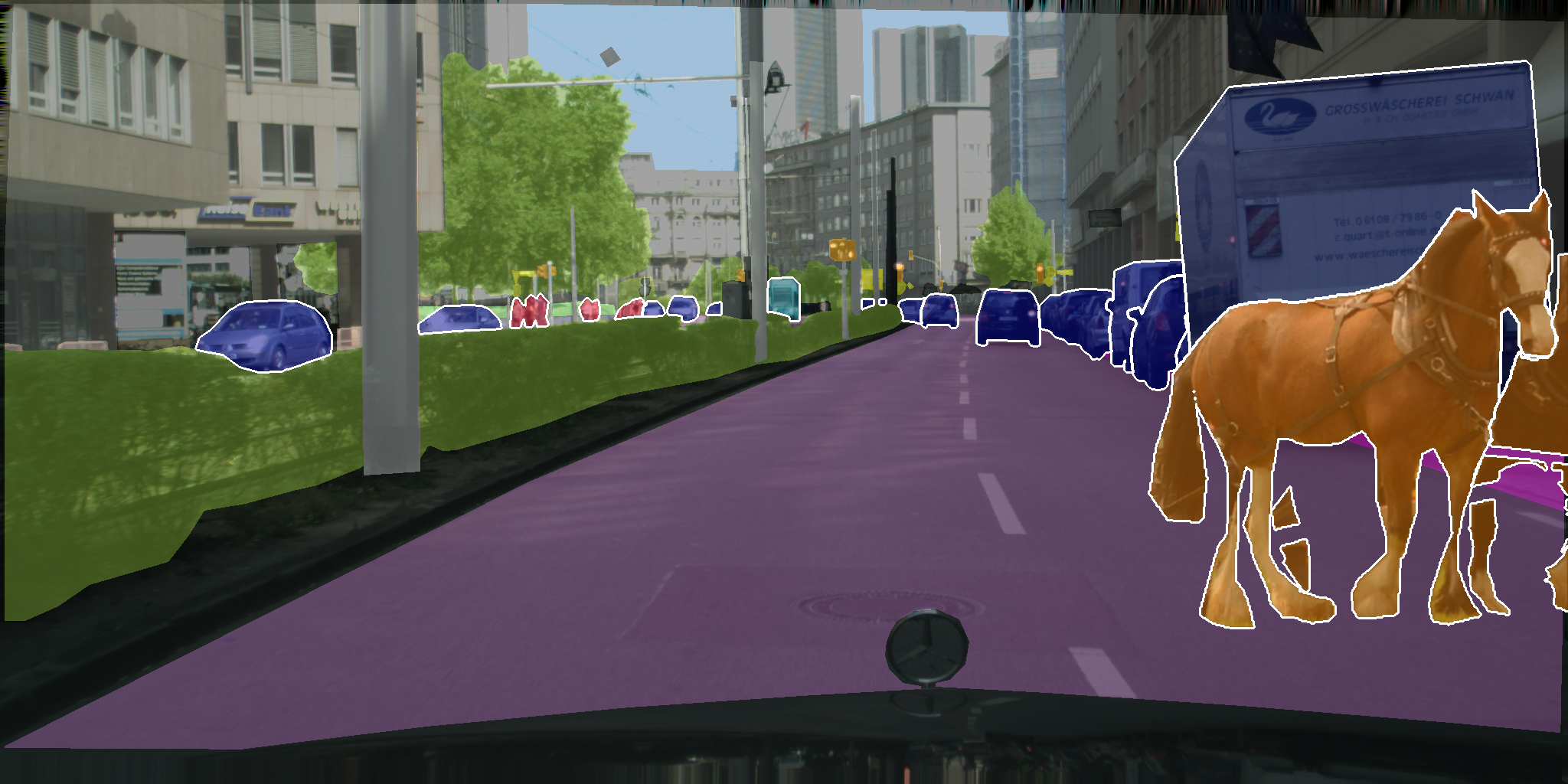}
        \subcaption{Depth-based scaling}
        \label{out depth-based }
    \end{subfigure}
    \caption{Incorporation of OOD objects into a scene from the Cityscapes dataset is shown, with (c) random scaling and (d) depth-based OOD object scaling.} 
    \label{fig:bins}
    \vspace{-0.4cm}
\end{figure}

\subsubsection{Datasets}
\label{subsubsec:dataset}

Given the high cost and complexity of annotating panoptic segmentation data, it is impractical to manually label a new dataset that encompasses a diverse set of real-world out-of-distribution instances for panoptic out-of-distribution segmentation. As an alternative, we extend established urban scene understanding datasets for panoptic segmentation by incorporating real-world OOD object instances, creating two new datasets: Cityscapes-OOD and BDD100K-OOD. 


{\parskip=3pt
\textit{Dataset Creation Protocol}: We extract atypical objects from the LVIS~\cite{gupta2019lvis} instance segmentation dataset using their segmentation masks. Objects such as cats and desks that are not present in the original panoptic segmentation dataset are added to the images. Their positions and the number of instances are randomized, with the object likelihood based on their typical locations (e.g., couches at the bottom, airplanes at the top). We further employ depth-dependent scaling to resize the OOD objects, ensuring that objects near the ego-car are relatively larger than the ones positioned far away. To do so, we begin by determining the sizes of objects in the original panoptic segmentation dataset and then match these sizes to bins established based on depth. Based on their size, the extracted OOD objects are paired with known semantic classes (e.g., surfboard with person, couch with car). We then overlay these objects, selecting a size from the depth bin randomly based on their positioning (\figref{fig:bins} (a) and (b)). We use blending techniques~\cite{blum2019fishyscapes} such as color shifts, depth blur, color curve, and gamma transformations, and we remove low-quality samples to enhance the dataset's quality and remove low-quality samples to improve the dataset's quality. Lastly, we ensure OOD objects in the training set are distinct from those in the test set, guaranteeing novelty during testing, and consistent with the requirements of the panoptic out-of-distribution segmentation task.\looseness=-1

\begin{figure}[t]
    \centering
    \begin{subfigure}[b]{0.49\linewidth}
        \centering
        \includegraphics[width=\textwidth]{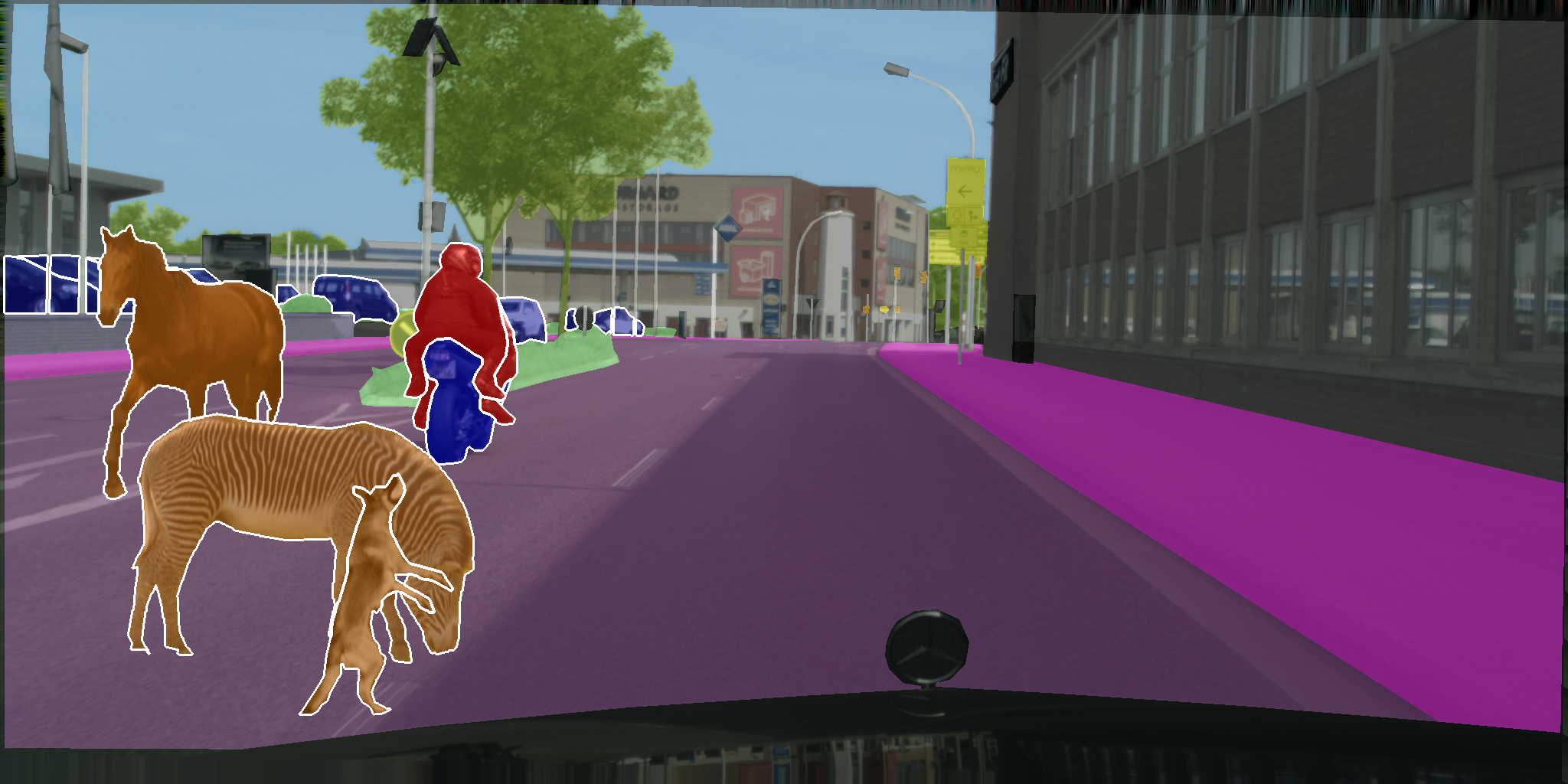}
        \subcaption{Cityscapes-OOD}
        \label{cityscapes_ex}
    \end{subfigure}
    \begin{subfigure}[b]{0.49\linewidth}
        \centering
        \includegraphics[width=\textwidth, trim={0 0 0 2.8cm},clip]{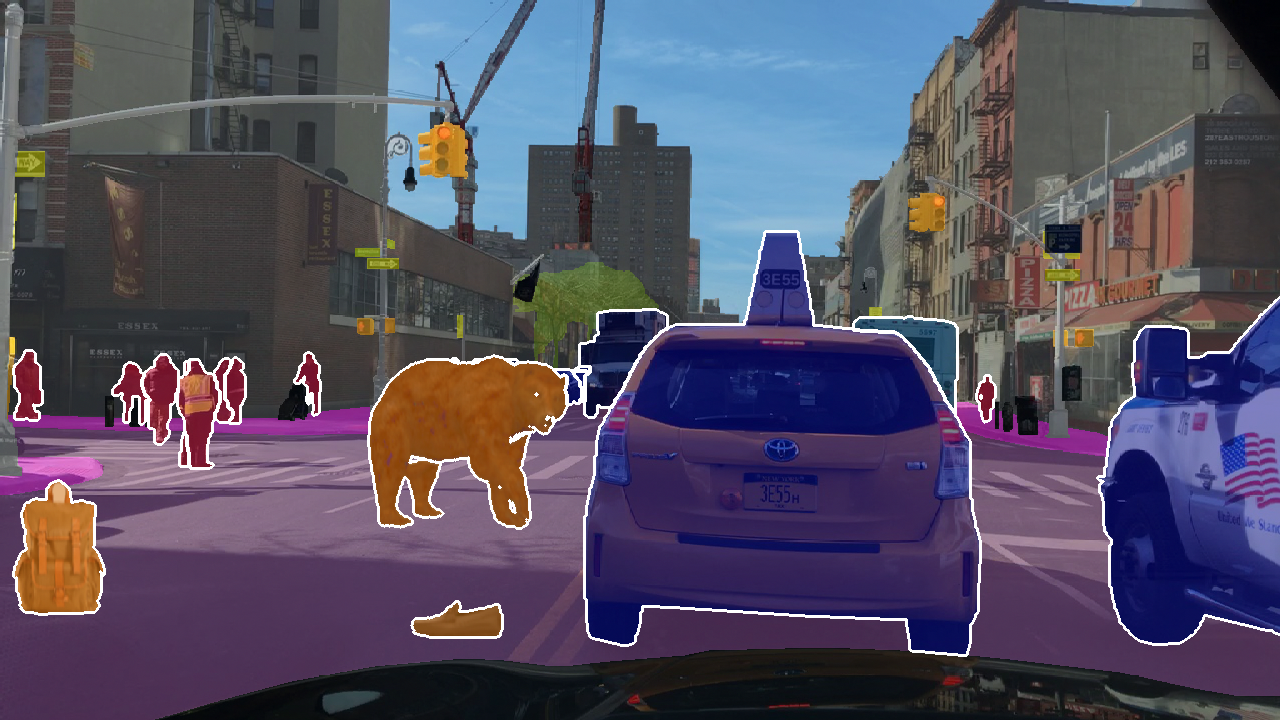}
        \subcaption{BDD100K-OOD}
        \label{cityscapes_ex}
    \end{subfigure}
\caption{Sample images from the Cityscapes-OOD and BDD100K datasets.}
    \label{fig:cityscapes-ood}
   \vspace{-0.3cm}
\end{figure}

\begin{figure}[t]
    \centering
    \begin{subfigure}[b]{0.49\linewidth}
        \centering
        \includegraphics[width=\textwidth]{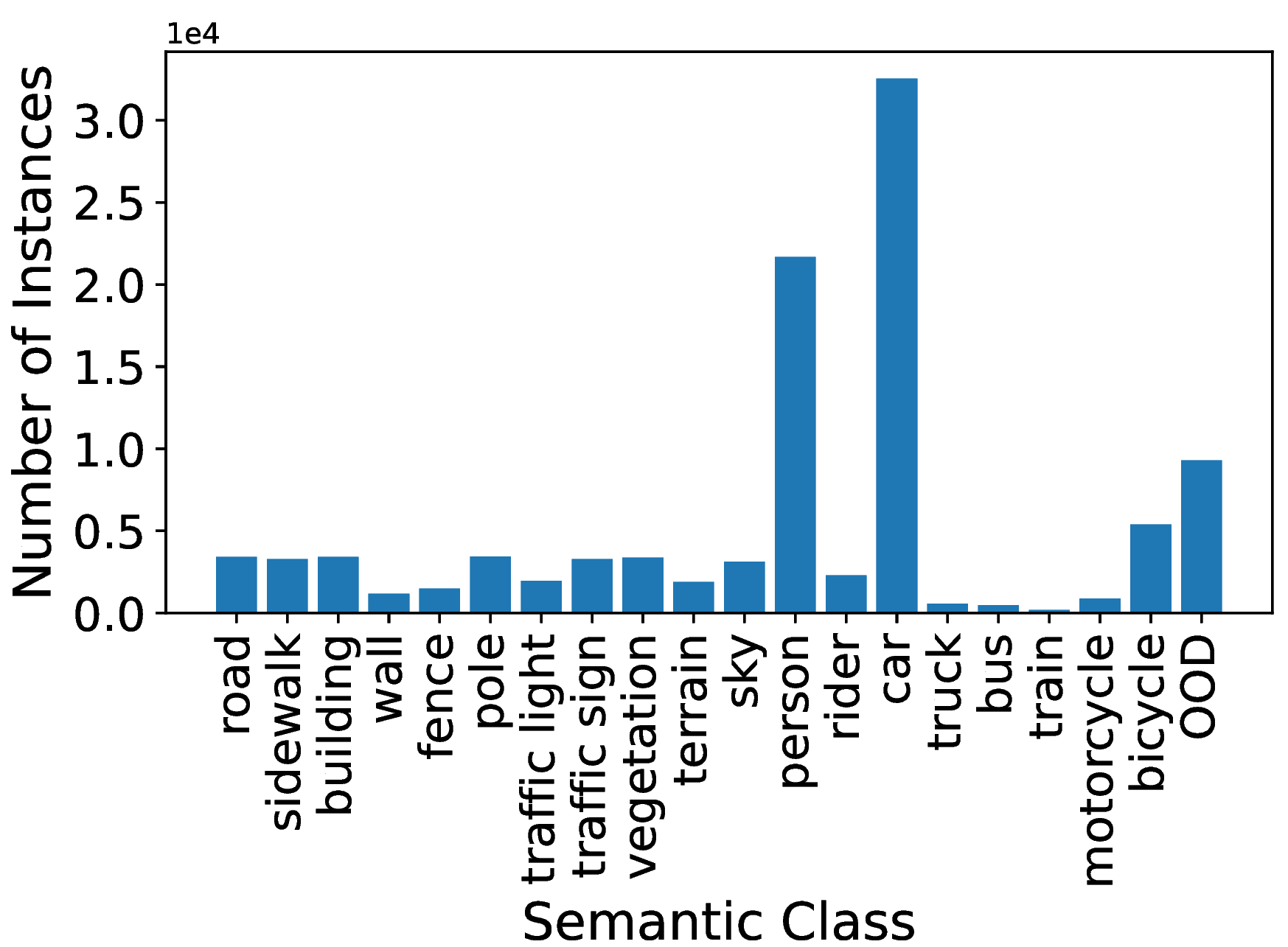} 
        \subcaption{Cityscapes-OOD dataset statistics.}
        \label{cityscapes_stats}
    \end{subfigure}
    \begin{subfigure}[b]{0.49\linewidth}
        \centering
        \includegraphics[width=\textwidth]{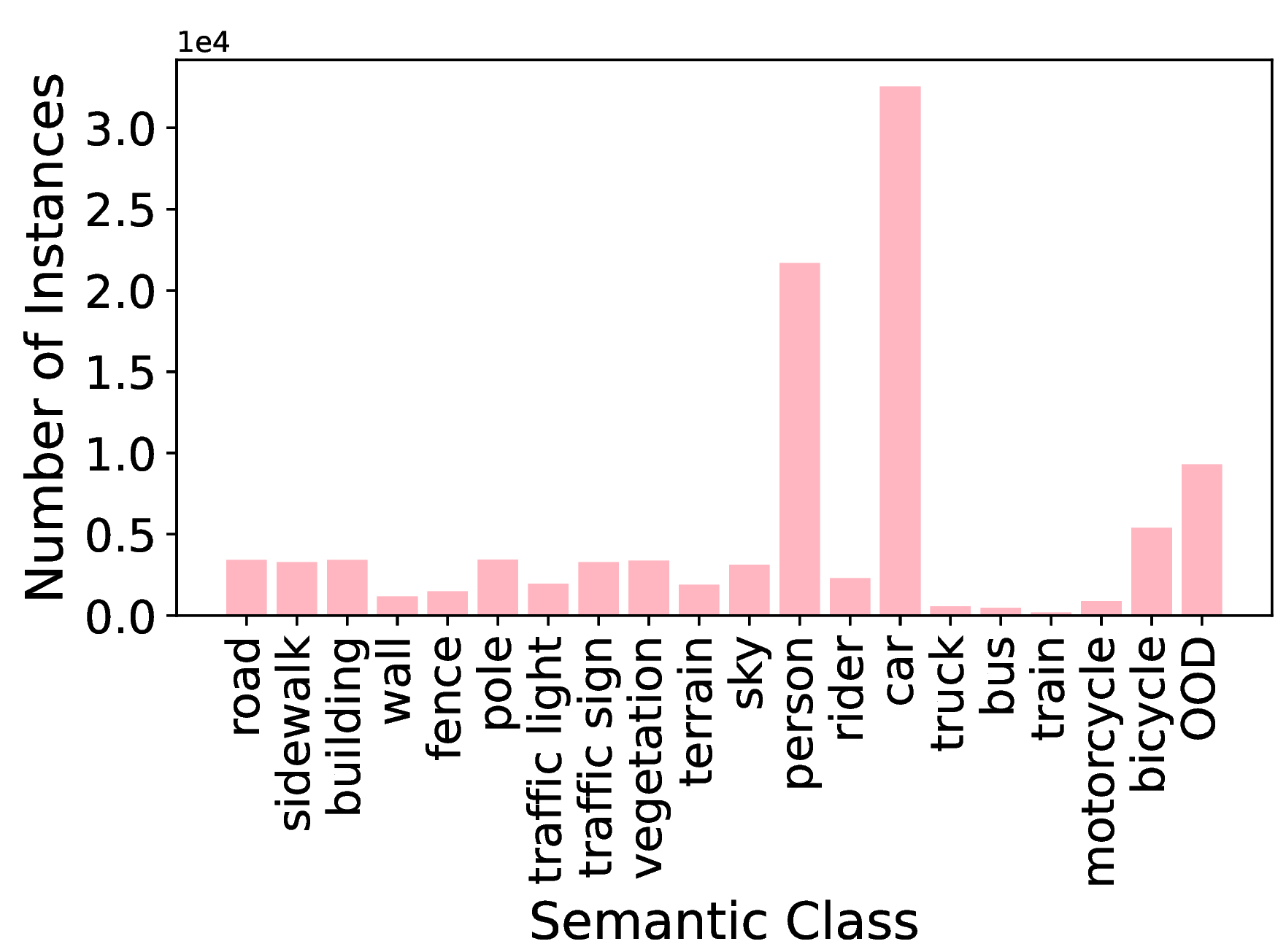}
        \subcaption{BDD100K-OOD dataset statstics.}
        \label{bdd_stats}
    \end{subfigure}
    \caption{Dataset statistics for (a) Cityscapes-OOD and (b) BDD100K-OOD. Note that each \textit{stuff} class has a single occurrence per image.}
    \label{fig:stats-ood}
    \vspace{-0.3cm}
\end{figure}

{\parskip=3pt
\textit{Cityscapes-OOD}: We create the Cityscapes-OOD dataset for panoptic out-of-distribution segmentation, with 11 \textit{stuff} classes, 8 \textit{thing} classes, and an \textit{OOD} class by extending Cityscapes~\cite{cordts2016cityscapes}. It consists of 2975 training and 500 test images at a resolution of $2048\times1024$ pixels. Test set annotations which are generated from the validation set of Cityscapes are not publicly released, and evaluation is only possible through an online server. \figref{fig:cityscapes-ood} (a) and \figref{fig:stats-ood} (a) shows an example and dataset statistics.

{\parskip=3pt
\textit{BDD100K-OOD} dataset consists of 7000 training and 1000 validation images with a resolution of $1280\times720$ pixels and is an extension of BDD100K~\cite{yu2020bdd100k}, augmented with out-of-distribution objects. It features one \textit{OOD} class, 11 \textit{stuff} classes including roads and buildings, and eight \textit{thing} classes such as cars and bicycles. \figref{fig:cityscapes-ood} (b) and \figref{fig:stats-ood} (b) present an example and dataset statistics, respectively.

\subsubsection{Baselines}
\label{subsubsec:baselines}

We present seven baselines for the panoptic out-of-distribution segmentation task. We adapt four effective semantic out-of-distribution segmentation methods (MSP~\cite{hendrycks2016baseline}, MaxLogit~\cite{hendrycks2019scaling}, ODIN~\cite{besnier2021triggering}, Meta-OOD~\cite{chan2021entropy}) with the PAPS~\cite{mohan22perceiving} modal panoptic segmentation architecture (PAPS$^*$ as described in~\secref{subsubsec:base}). We compute the OOD semantic class from the semantic segmentation output from PAPS$^*$ and then use the post-processing approach from \cite{cheng2020panoptic} for \textit{thing}$+$\textit{OOD} foreground segmentation to obtain the final panoptic out-of-distribution segmentation prediction. For the two remaining baselines, EPSON~\cite{hwang2021exemplar} and DD-OPS~\cite{xu2022dual}, we restrict the segmentation of unknown classes to a single \textit{OOD} class and enhance their base network with EfficientPS~\cite{mohan2020efficientps}, a state-of-the-art top-down panoptic segmentation network.



\section{PoDS Network Architecture}\label{sec:method}
\begin{figure*}
    \centering
    \begin{subfigure}[b]{\linewidth}
        \centering
        \includegraphics[width=0.92\linewidth]{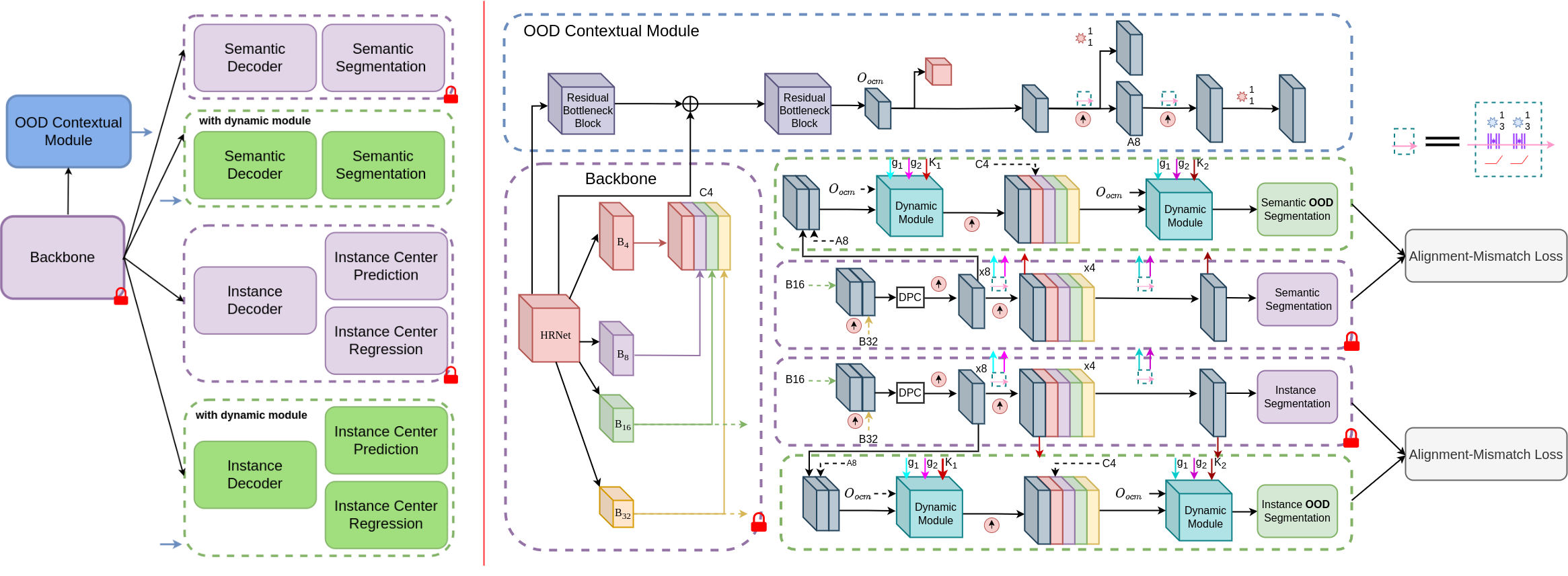}
        \label{APSNEt_arch}
    \end{subfigure}
        \hfill
    \begin{subfigure}[b]{0.6\linewidth}
       \centering
        \includegraphics[width=\textwidth]{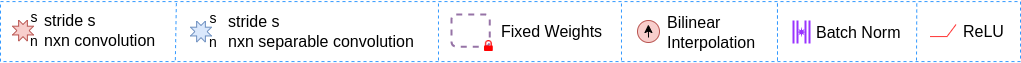}
    \end{subfigure}
    \caption{Illustration of our proposed PoDS architecture that consists of a shared backbone with an OOD contextual module and symmetrical task-specific decoder arranged in a dual configuration setup to facilitate an alignment-mismatch learning strategy. The shared backbone learns robust feature representations for in-distribution semantic categories while the OOD contextual module supports both global and local features for OOD objects. The network comprises symmetrical semantic and instance decoders that include dynamic modules to adaptively balance the features between in- and out-distribution representations.}
    \label{fig:network}
    \vspace{-0.3cm}
\end{figure*}

In this section, we detail our proposed PoDS architecture depicted in~\figref{fig:network}. We first present an overview of the PoDS network, followed by a detailed description of each constituting component. PoDS builds on top of a base panoptic segmentation network that has a shared backbone and task-specific decoders (purple) by incorporating modules specially designed to embed out-of-distribution capabilities based on prior knowledge of in-distribution classes. We incorporate an OOD contextual module (blue) that complements the robust in-distribution semantic features of the shared backbone with both global discriminatory and fine local OOD object representations. Subsequently, we introduce an additional task-specific decoder (green), equipped with dynamic modules, alongside the existing ones. This design allows for adaptive integration of OOD features while preserving the in-distribution features of the high-performing base panoptic network. The unique dual task-specific decoder configuration benefits further from our novel alignment-mismatch loss. This loss encourages learning finer details within in-distribution semantic classes and what lies outside by balancing consensus and divergence between the two heads. Furthermore, we incorporate a data augmentation strategy to facilitate the training of our novel modules. Please refer to \secref{sec:sarch} of the supplementary material for further implementation details.

\subsubsection{Base Network}
\label{subsubsec:base}

Building on the modal variant of PAPS~\cite{mohan22perceiving}, which excels in panoptic segmentation, we develop an architecture for panoptic out-of-distribution segmentation. The modal PAPS architecture has a shared backbone, decoders, prediction heads, a context extractor, and a cross-task module. For our PoDS network, we streamline this by excluding the instance segmentation decoder and cross-task module. Instead, we adopt the semantic segmentation decoder with the dense prediction cell module, along with two upsampling stages and skip connections for the instance segmentation decoder. The instance segmentation head remains intact, handling instance center prediction and regression. This streamlined PAPS architecture, termed PAPS$^*$, achieves a PQ score of $63.7$ on the Cityscapes validation set, close to PAPS's $64.3$. As shown in~\figref{fig:network} purple boxes with red locks, we pretrain PAPS$^*$ on in-distribution panoptic segmentation datasets, and keep its weights fixed throughout the out-of-distribution segmentation training.

\subsubsection{OOD Contextual Module} 
\label{subsubsec:GCS}
We introduce the OOD Contextual Module for the PoDS architecture, designed to capture both global and local features of out-of-distribution (OOD) objects in images. As depicted in \figref{fig:network} (blue box), this module incorporates two residual bottleneck blocks, similar to the fourth and fifth stages of Regnet~\cite{xu2022regnet}. The module takes the output from the last layer of the backbone's stage 2, processes it through the first block, combines it with the output from the last layer of stage 3, and routes it through the second block. Subsequently, the output from the second block, named O\textsubscript{ocm}, proceeds to a global average pooling layer and then to a classification head. In parallel, O\textsubscript{ocm} undergoes upsampling followed by two convolution layers. This processed output splits into two branches: one path goes to an in/out-distribution segmentation head, while the other undergoes further upsampling and convolutions before reaching a second segmentation head. Both heads distinguish between pixel-wise in-distribution and out-of-distribution regions. During training, we apply random data augmentation, as detailed in~\secref{subsubsec:data_aug}, generating samples with OOD objects from the panoptic segmentation dataset to train both the classification (targeting OOD global features) and pixel-wise segmentation heads (focusing on OOD local features). We employ binary cross-entropy loss for the classification head and weighted pixel-wise cross-entropy loss for the in/out-distribution segmentation heads, with weights of 0.7 and 0.8 respectively. Thus, $L_{ocm}$ is the sum of the aforementioned losses. Notably, backpropagation for the pixel-wise loss is only triggered when an OOD object is present in the input.

\subsubsection{Dynamic Module} 
\label{subsubsec:dynamic}



The dynamic module is defined by the following inputs: an input feature map $F$, $O_{ocm}$ ($G$) from the OOD contextual module, and feature map and convolution functions ($K$ and $g_1(\cdot, w_1)$ and $g_2(\cdot, w_2)$, respectively) from the base network. The inputs from the base network are represented by the red, yellow, and pink arrows in \figref{fig:network}, respectively.
\begin{align}
    F_{R} &= g_1^*(F, w_1 + \Delta w_1^*) , \\  
    F_{R} &= g_2^*(F_{R}, w_2 + \Delta w_2^*) , \\  
    F_{O} &= h_1(G)\cdot F_{R} + (1-h_1(G))\cdot K,
\end{align}
where $g_1^*(\cdot, w_1 + \Delta w_1^*)$ and $g_2^*(\cdot, w_2 + \Delta w_2^*)$ denote convolution functions with learned weights $w^*$ as offsets from the weights $w$ of $g_1(\cdot, w_1)$ and $g_2(\cdot, w_2)$, respectively. The computation of $F_{R}$ is performed in a sequential manner with an initial input $F$ using the functions $g_1^*(\cdot)$ and $g_2^*(\cdot)$. The weighted gating function $h_1$ is composed of a consecutive global pooling and a $1 \times 1$ convolution layer. By using offset weights and OOD contextual features O\textsubscript{ocm}, the proposed module establishes a pre-training and training link between the convolutional weights of the base network and its own weights, which can be dynamically adjusted. This allows the module to maintain the pre-learned knowledge of the base network while incorporating out-of-distribution capabilities, making minor adjustments if the OOD object is similar to known semantic classes and significant adjustments if it is considerably different.

\subsubsection{PoDS Decoders and Heads}
In the PoDS framework, we utilize additional decoders akin to the base network described in \secref{subsubsec:base}, as visualized with green boxes in \figref{fig:network}. Each decoder starts by merging the upsampled DPC features from their base network's task-specific decoders (purple boxes) with the $A_8$ features from the OOD contextual module (blue boxes). The resulting features ($F$) are then processed by a dynamic module, which also takes in the output of the existing convolution layers ($K_1$) of the $\times8$ stage in the base network. The output of the module is then upsampled and concatenated with $C_4$ (\secref{subsubsec:base}) and both are fed to another dynamic module along with the output of the existing convolution layers ($K_2$) of $\times4$ stage in the base network. The final output of this module is then fed to the corresponding task-specific heads.

The PoDS base network targets only in-distribution classes. To expand the network's capabilities, we incorporate additional task-specific heads that can learn about both in-distribution and out-of-distribution classes. These heads consist of two sequential layers of $3\times3$ depthwise-separable convolutions, followed by a task-specific $1\times1$ predictor. The \textit{OOD} semantic segmentation head uses a predictor with $N_{stuff}+N_{thing}+1$ for segmentation labels. The \textit{OOD} instance segmentation heads have two predictors: instance center prediction and instance center regression, which learn on \textit{thing} + \textit{void} regions. To train the semantic head, we use the weighted bootstrapped cross-entropy loss ($L_{sem}$). For the instance center prediction, we use the Mean Squared Error (MSE) loss ($L_{cp}$) to minimize the distance between the predicted heatmaps and the 2D Gaussian encoded groundtruth heatmaps. For instance center regression, we use the $L1$ loss.

\subsubsection{Learning from alignment-mismatch}
\label{subsubsec:conflict}
We train the semantic segmentation head $SH_{in}$ from the PoDS base network only for known in-distribution classes and we train the PoDS head $SH_{out}$ for an added \textit{OOD} class. The $SH_{in}$ consistently labels pixels with known semantic classes, irrespective of in- or out-of-distribution object. During the training of $SH_{out}$, we aim to amplify the prediction discrepancies between $(SH_{in}$ and $SH_{out})$ for out-of-distribution class pixels while promoting consensus for in-distribution object predictions. 
To implement the alignment-mismatch strategy, we ensure the output dimensions of $SH_{in}$ and $SH_{out}$ match. Given $SH_{in}$  has $(N_{stuff}+N_{thing})\times H\times W$ channels and $SH_{out}$ has  $(N_{stuff}+N_{thing}+1)\times H\times W$, we derive an extra channel for $SH_{in}$ by taking the maximum across the semantic class dimension. 
We employ the following loss to foster alignment-mismatch between the two heads, depending on whether the pixel belongs to an in-distribution or out-of-distribution object:
\begin{align}
    e_i &= |s(SH_{in}(x_i)) - s(SH_{out}(x_i))|^2, \\
    s_i &= (1 - y_i) e_i, \label{eq:1} \\
    d_i &= y_i \max (0, m - e_i), \label{eq:2} \\
    L_{s-am} &= \frac{1}{2N} \sum_{i=1}^{N} s_i + d_i,
\end{align}
where $N$ is the number of pixels, $x_i$ is the input image pixel, $y_i$ is the label that indicates out- or in-distribution class, $m$ is the hyperameter and $s = \ln\left(1 + \text{e}^x\right)$ is the softplus acitvation function. We use the softplus to encourage that $SH_{out}$ predicts void class for out-of-distribution pixels with higher logits compared to $SH_{in}$'s maximum logit. We use softplus to ensure $SH_{out}$ predicts the ood class for out-of-distribution pixels with logits higher than the maximum logit from $SH_{in}$. Since softplus always yields positive output and weights of $SH_{in}$ are frozen, the only way for $SH_{out}$ to reduce the loss is by predicting out-of-distribution classes with high logits, especially when the margin hyperparameter m in the loss is sufficiently large.



For instance segmentation, we strive to foster alignment-mismatch between the instance center prediction and instance center regression heads. We achieve this by employing a similar loss as $L_{s-am}$ but applied in the feature space to the features $X_{in}$ and $X_{out}$ prior to the predictor for respective heads. This separates in-distribution from out-of-distribution features, making it easier to perform the center prediction and center regression. We define instance segmentation losses as
\begin{align}
    e_i &= |X_{j-in}^i - X_{j-out}^i|, \\
    L_{j-am} &= \frac{1}{2N} \sum_{i=1}^{N} s_i + d_i,
\end{align}
where $X_{j-in}^i$ and $X_{j-out}^i$ are the features computed at location $i$ for the instance segmentation heads. $s_i$ and $d_i$ are the same as \eqref{eq:1} and \eqref{eq:2}, respectively. $j \in [c,r]$ represents the instance center prediction or instance center regression heads, from which we obtain the losses $L_{c-am}$ and $L_{r-am}$, respectively.

\subsubsection{Data Augmentation}
\label{subsubsec:data_aug}
For training the PoDS architecture, we mix samples containing solely in-distribution classes with those that include both in- and out-of-distribution objects. To curate out-of-distribution samples, we source web images via specific keywords, ensuring they exclude known in- or out-of-distribution objects from the Cityscapes-OOD and BDD100K-OOD test set. Using an unsupervised instance segmentation network~\cite{wang2022freesolo}, we generate pseudo instance masks for these images, facilitating the extraction and compilation of a diverse out-of-distribution (OOD) object repository. During training, images are either augmented with randomly positioned and scaled OOD objects or left as in-distribution. Additionally, the compiled OOD objects are split into two types based on similarity to known semantic classes. Initially, training focuses on vastly dissimilar objects such as hair dryers, with a progressive shift towards more similar objects such as monkeys, ensuring gradual learning from distinct to closely related OOD objects.\looseness=-1

\section{Experimental Evaluation}\label{sec:experiments}

\begin{table}
\setlength\tabcolsep{3.0pt}
\centering
\caption{Panoptic out-of-distribution benchmarking results on the Cityscapes-OOD and BDD100K-OOD test set. 
Subscripts $out$ and $in$ refer to out-of-distribution class and in-distribution classes, respectively. All scores in [\%].}
\footnotesize
\begin{tabular}{l|ccc|ccc} 
\toprule
Model &  \multicolumn{3}{c|}{Cityscapes-OOD} & \multicolumn{3}{c}{BDD100K-OOD} \\ 
\cmidrule{2-7}
 &  POD-Q  & PQ$_{out}$ & PQ$_{in}$&  POD-Q  & PQ$_{out}$ & PQ$_{in}$ \\
\midrule
MSP~\cite{hendrycks2016baseline}  & $12.8$ & $3.4$ & $47.6$ & $9.1$ & $2.6$ & $32.1$\\
MaxLogit~\cite{hendrycks2019scaling} & $15.9$ &  $5.2$ & $48.6$ & $12.7$ & $4.7$ & $34.5$\\
ODIN~\cite{liang2017enhancing}   & $20.8$ &  $8.7$ & $49.8$ & $16.9$ & $7.9$ & $36.1$\\
EPSON~\cite{hwang2021exemplar}   & $29.4$ &  $15.9$ & $54.4$ & $23.7$ & $14.3$ & $39.4$\\
Meta-OOD~\cite{chan2021entropy}   & $41.7$ &  $31.3$ & $55.6$ & $34.5$ & $28.6$ & $41.6$\\ 
DD-OPS~\cite{xu2022dual}   & $46.1$ & $36.1$ & $58.7$ & $38.0$ & $33.2$ & $43.5$\\
\midrule
\net (Ours) &  $\mathbf{53.4}$ &  $\mathbf{45.9}$  & $\mathbf{62.2}$ & $\mathbf{42.3}$ & $\mathbf{38.7}$ & $\mathbf{46.3}$\\ 
\bottomrule
\end{tabular}
\label{tab:cityOODEvaluation}
\vspace{-0.3cm}
\end{table}

\subsection{Training and Inference Protocol}
\label{subsubsec:training}

We adopt a two-stage training approach for our network. Initially, we train the base layers of the PoDS network for 160,000 iterations on Cityscapes and 240,000 on BDD100K to instill strong in-distribution priors. Subsequently, these base layers are frozen and the other components of the PoDS network are trained using the data augmentation techniques outlined in \secref{subsubsec:data_aug} for 90K and 150K iterations on Cityscapes and BDD100K, respectively. For each training phase, we employ the Adam optimizer with a poly learning rate schedule, setting the initial learning rates at $0.001$ for Cityscapes and $0.005$ for BDD100K. We optimize the following loss functions for training the network:
\begin{align}
\begin{split}
    L &= L_{ocm} + L_{sem} + \alpha L_{cp} \\
    & \qquad + \beta_{1}(L_{cr} + L_{s-ad})  + \beta_{2}({L_{c-ad} + L_{r-ad}}),
\end{split}
\end{align}
where $\alpha=200$, $\beta_{1}=0.01$ and $\beta_{2}=0.001$. All of the individual losses are defined in \secref{sec:method}. We set the margin hyperparameter $m$ to 50. During inference, we use the same post-processing as~\cite{cheng2020panoptic} with the semantic and instance segmentation head predictors that learn with the inclusion of OOD class.

\subsection{Benchmarking Results}
\label{subsubsec:benchmarking}
In \tabref{tab:cityOODEvaluation}, we compare the performance of our PoDS architecture with the baselines on the Cityscapes-OOD and BDD100K-OOD test sets. The first three baselines, MSP~\cite{hendrycks2016baseline}, MaxLogit~\cite{hendrycks2019scaling}, and ODIN~\cite{liang2017enhancing}, adapt any panoptic segmentation network for out-of-distribution segmentation without modifications. We observe that they perform poorly compared to other reported methods as the task also requires identifying instances of out-of-distribution objects and not only obtaining dense predictions for them. While thresholding confidence scores from these baselines enhances OOD object sensitivity, it often results in fragmented detections and misclassifications of in-distribution objects as OOD. Consequently, these baselines are not ideal for directly employing them for panoptic out-of-distribution segmentation.
EPSON~\cite{hwang2021exemplar} mines labels from void regions to learn clusters for OOD objects. While it improves OOD detection and reduces in-distribution misclassification, its low POD-Q scores indicate limited generalization to unseen \textit{OOD} objects during testing. Meta-OOD~\cite{chan2021entropy} emphasizes higher entropy for OOD predictions, whereas DD-OPS~\cite{xu2022dual} refines the utilization of void regions, rejecting objects using known classes and employing a class-agnostic classifier for OOD determination. PoDS employs a dual-head predictive setting to delineate the boundaries between in-distribution and out-of-distribution classes and to increase confidence in OOD object segmentation during training. By leveraging the alignment and mismatch between the heads, PoDS achieves the highest POD-Q scores of $53.4$ on Cityscapes-OOD and $42.3$ on BDD100K-OOD.

\begin{figure*}
\centering
\footnotesize
{\renewcommand{\arraystretch}{0.5}
\begin{tabular}{P{0.4cm}P{3.8cm}P{3.8cm}P{3.8cm}P{3.8cm}}
&  \raisebox{-0.4\height}{Image} &  \raisebox{-0.4\height}{Ground Truth} & 
\raisebox{-0.4\height}{DD-OPS~\cite{xu2022dual}} & \raisebox{-0.4\height}{PoDS (Ours)}\\
\\
\rot{(a)} 
& \raisebox{-0.4\height}{\includegraphics[width=\linewidth]{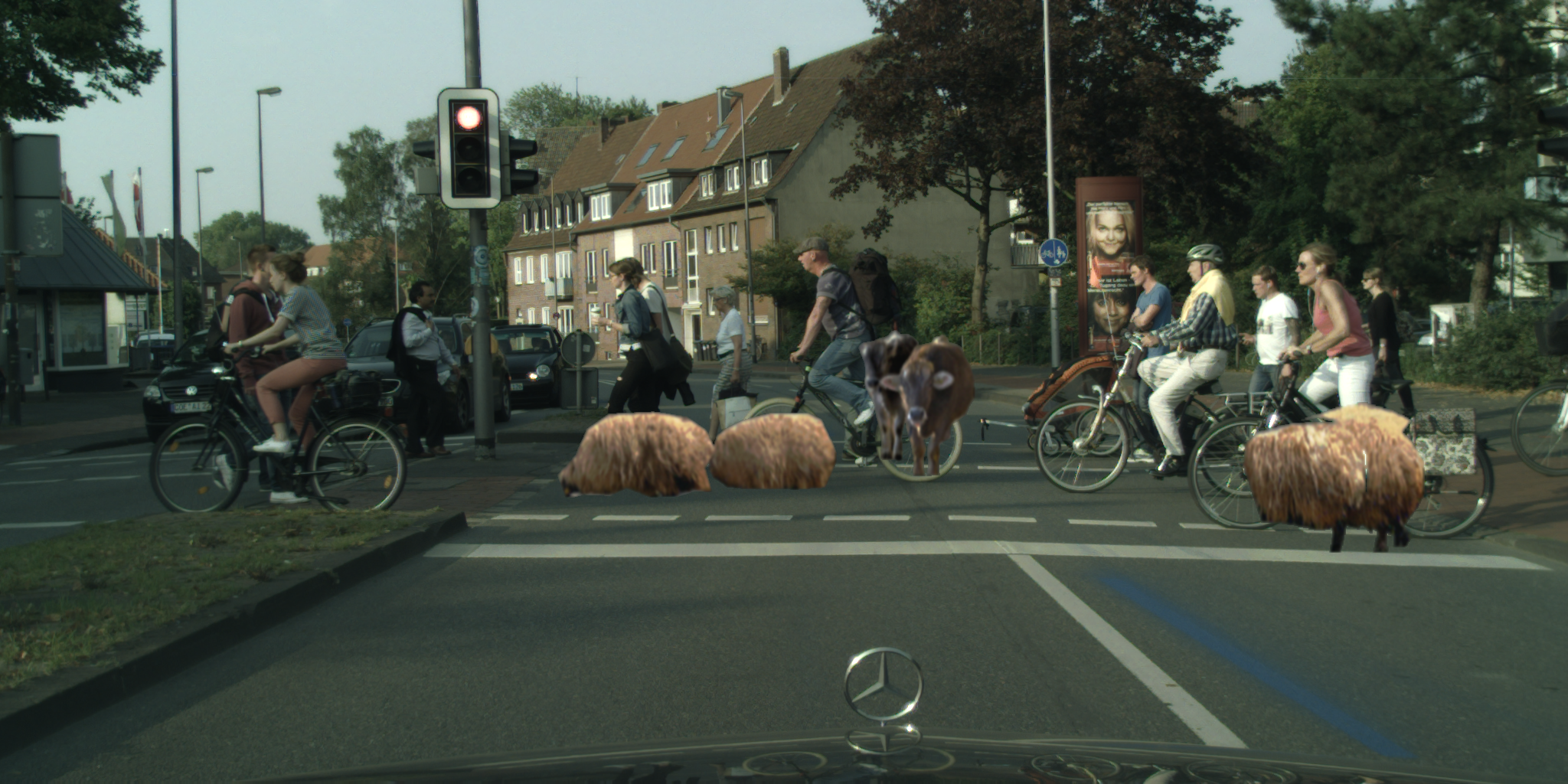}} & \raisebox{-0.4\height}{\includegraphics[width=\linewidth]{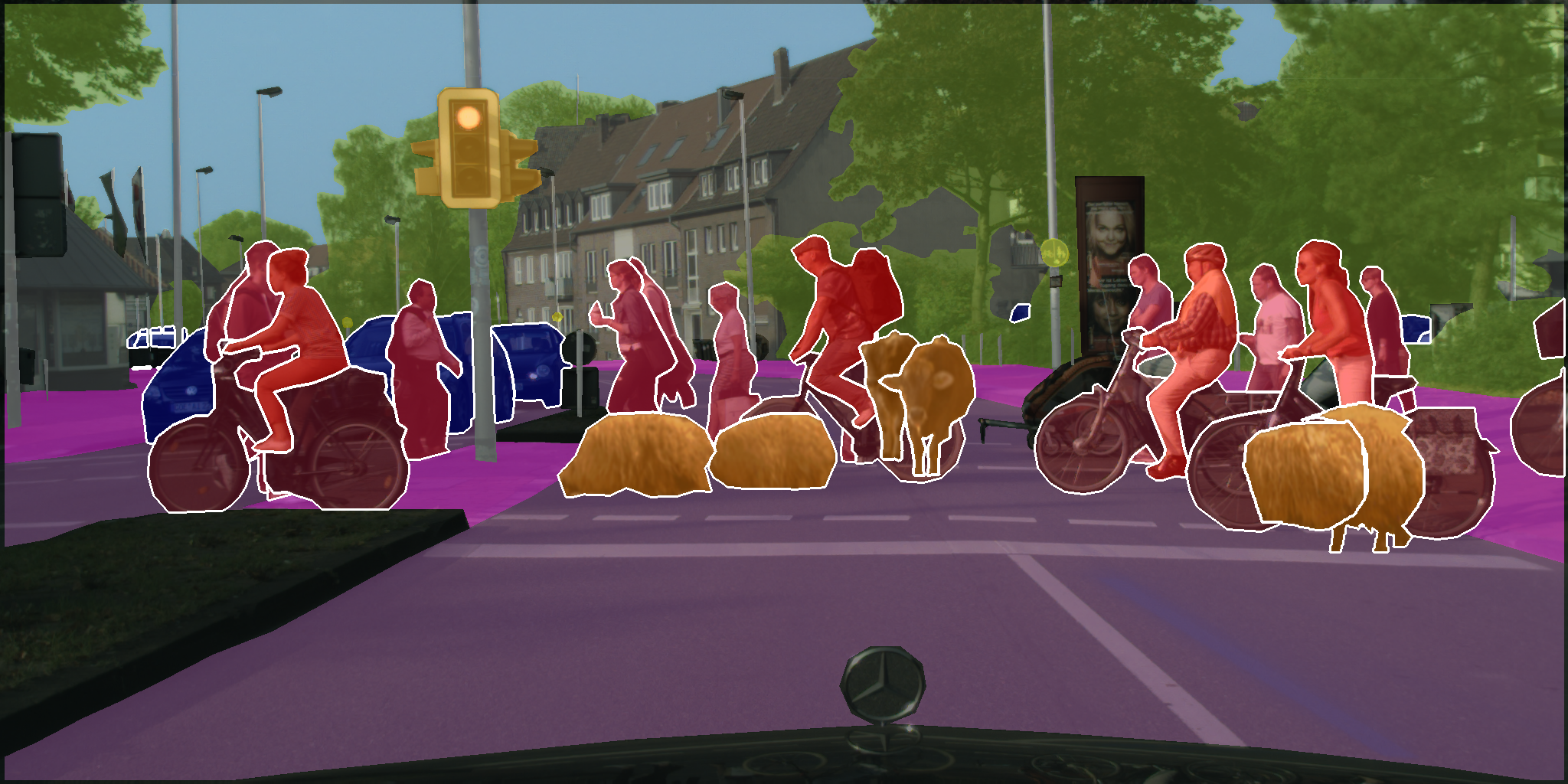}} & \raisebox{-0.4\height}{\includegraphics[width=\linewidth,frame]{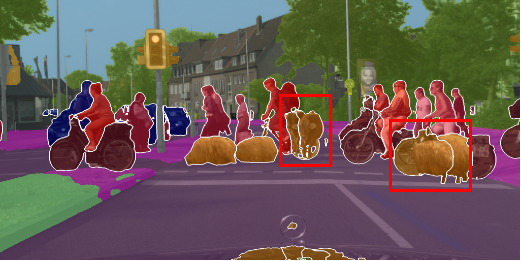}} &  \raisebox{-0.4\height}{\includegraphics[width=\linewidth]{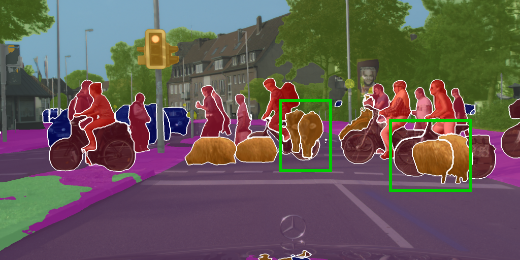}}\\
\\

\rot{(b)} 
& \raisebox{-0.4\height}{\includegraphics[width=\linewidth]{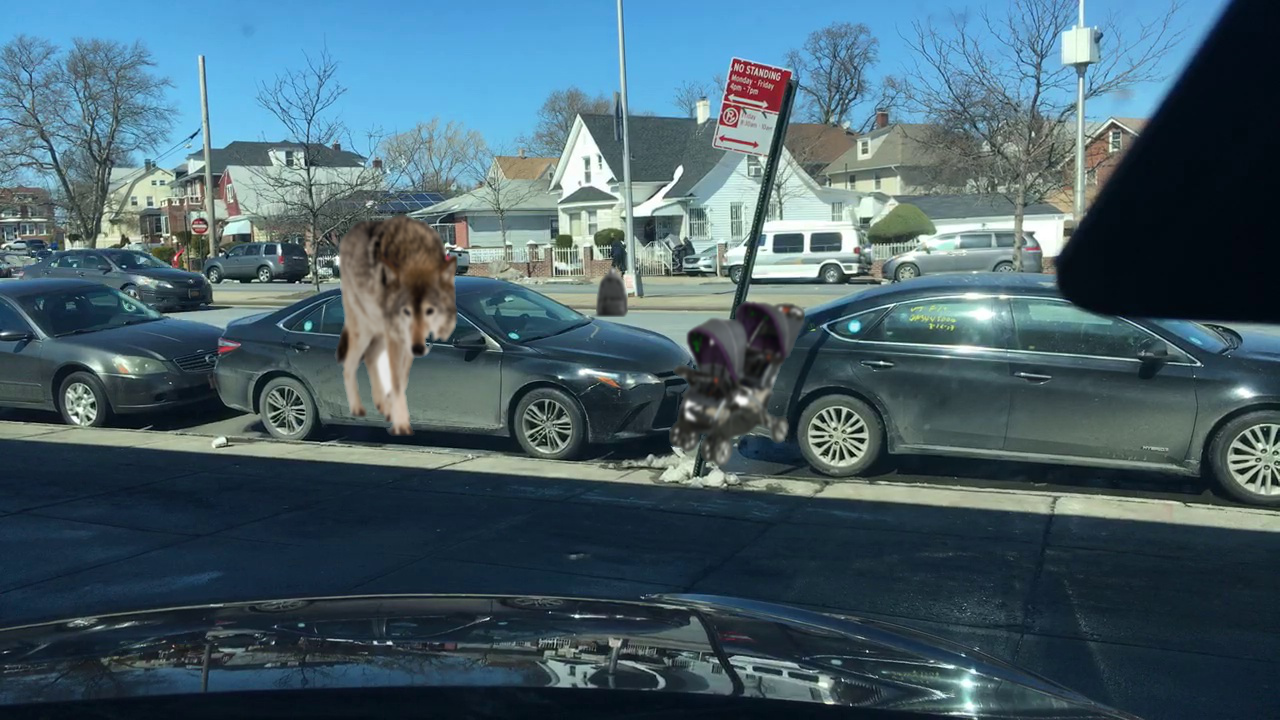}} & \raisebox{-0.4\height}{\includegraphics[width=\linewidth]{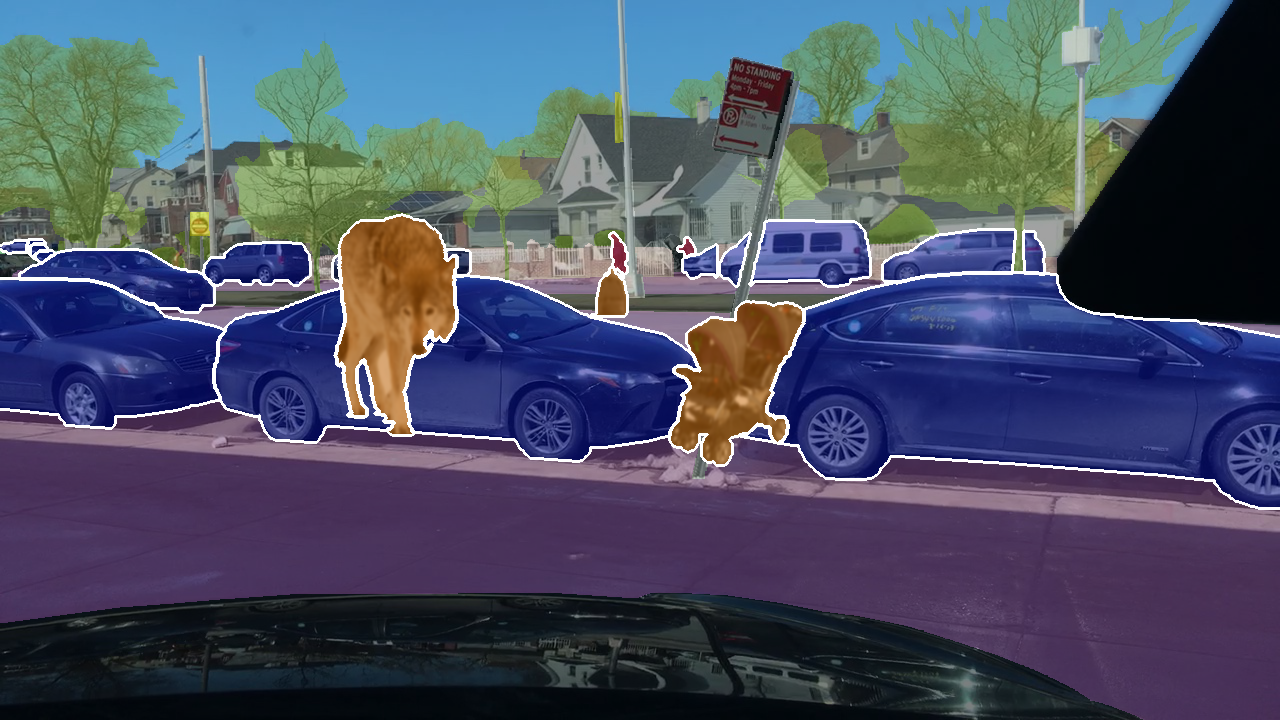}} & \raisebox{-0.4\height}{\includegraphics[width=\linewidth,frame]{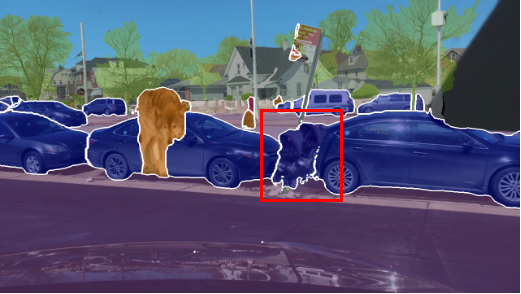}} &  \raisebox{-0.4\height}{\includegraphics[width=\linewidth]{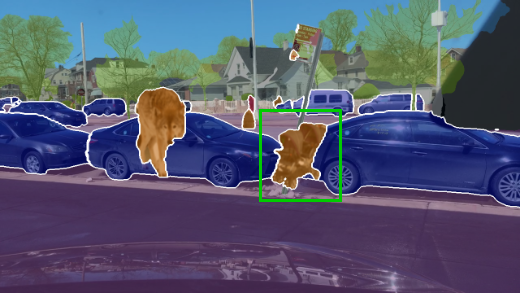}}\\
\\
\rot{(c)} 
& \raisebox{-0.4\height}{\includegraphics[width=\linewidth]{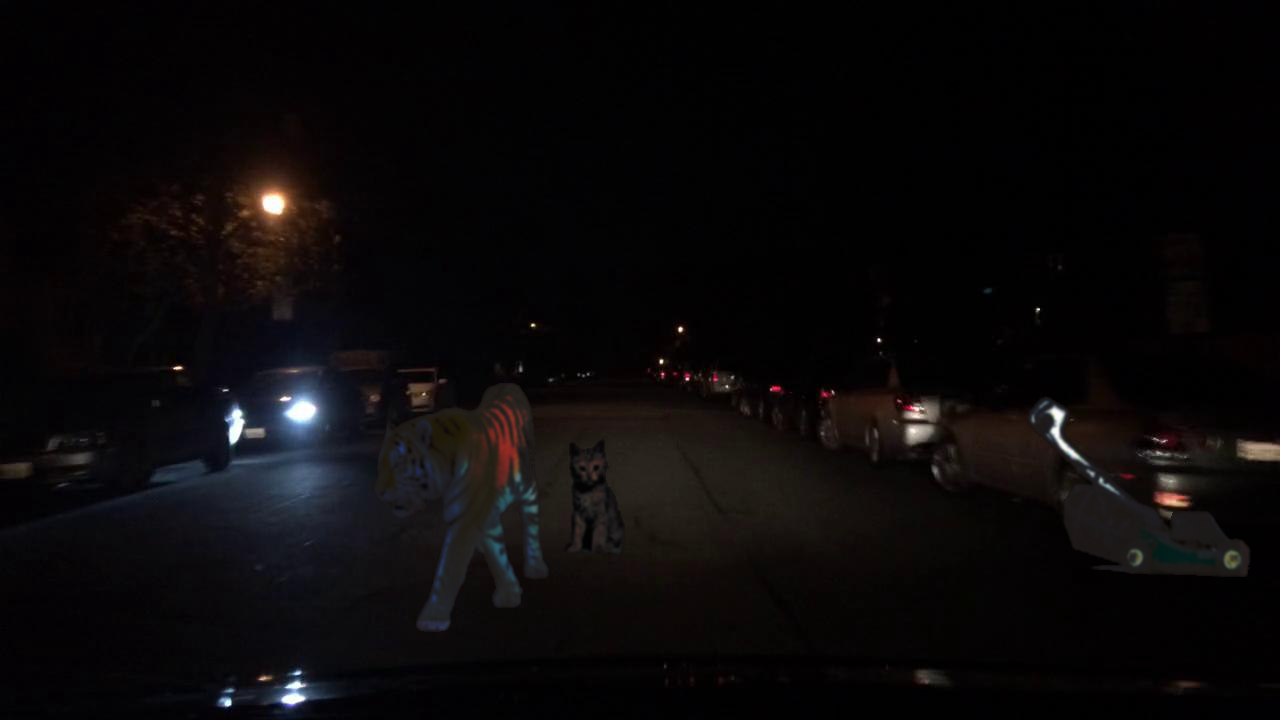}} & \raisebox{-0.4\height}{\includegraphics[width=\linewidth]{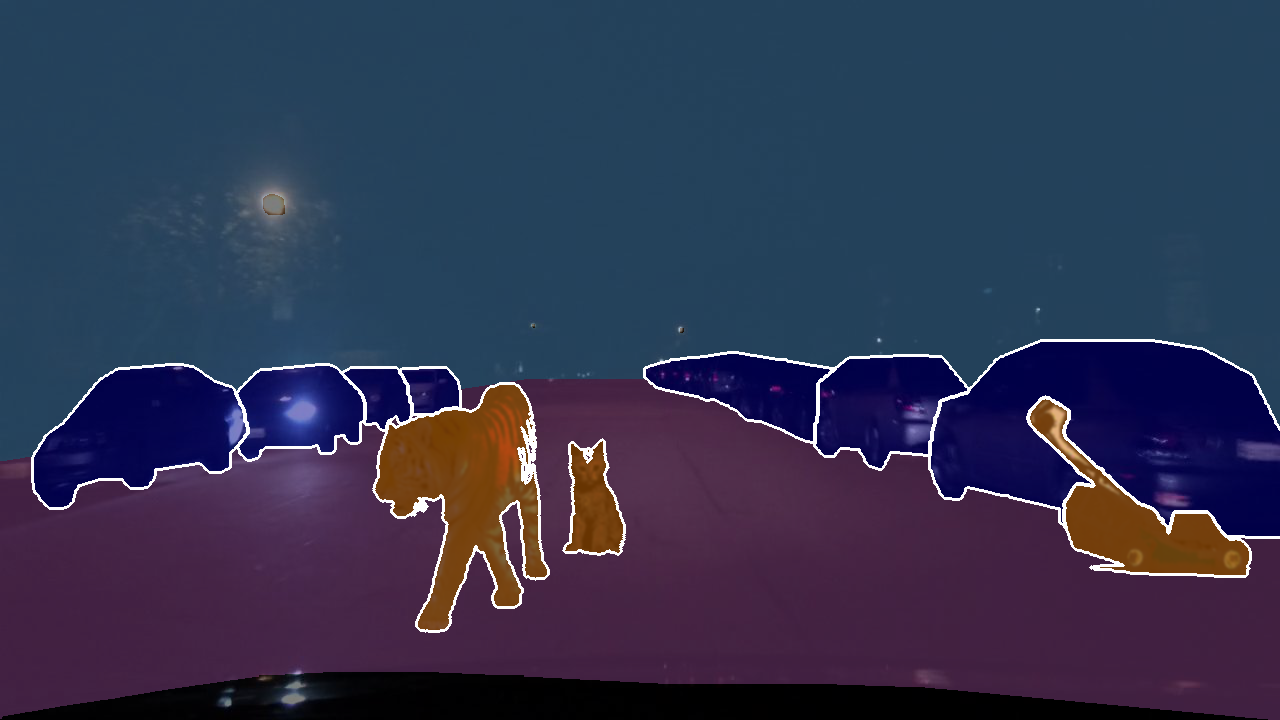}} & \raisebox{-0.4\height}{\includegraphics[width=\linewidth,frame]{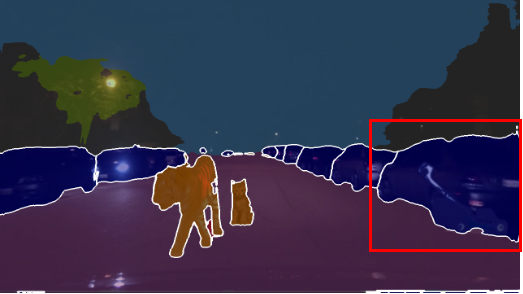}} &  \raisebox{-0.4\height}{\includegraphics[width=\linewidth]{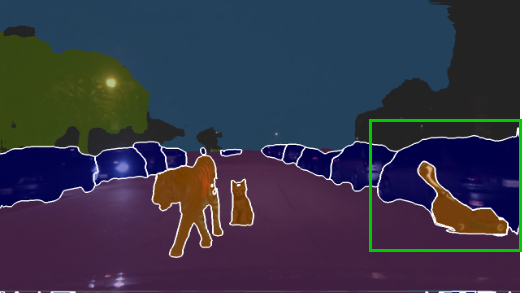}}\\
\\

\rot{(d)} 
& \raisebox{-0.4\height}{\includegraphics[width=\linewidth]{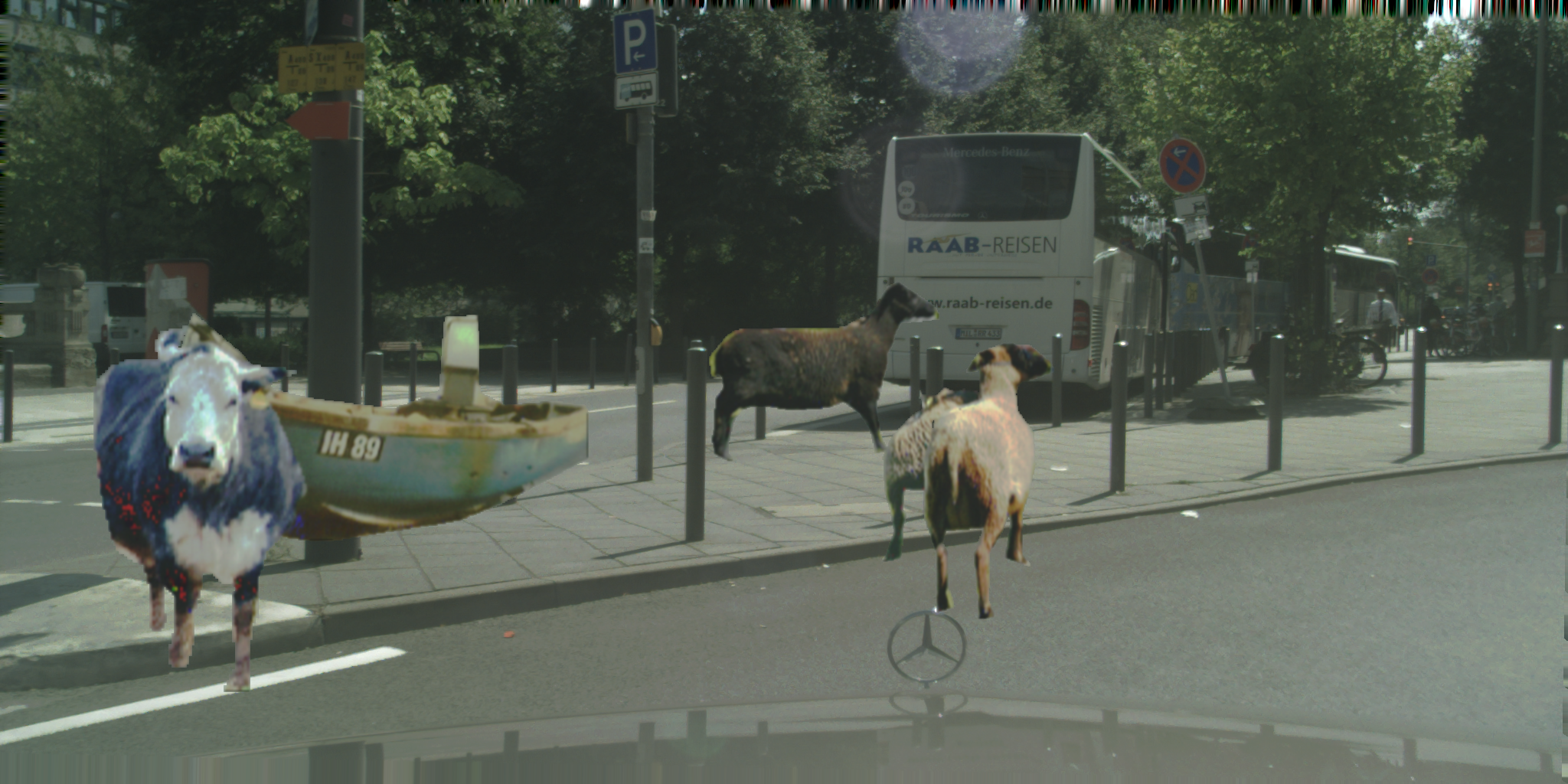}} & \raisebox{-0.4\height}{\includegraphics[width=\linewidth]{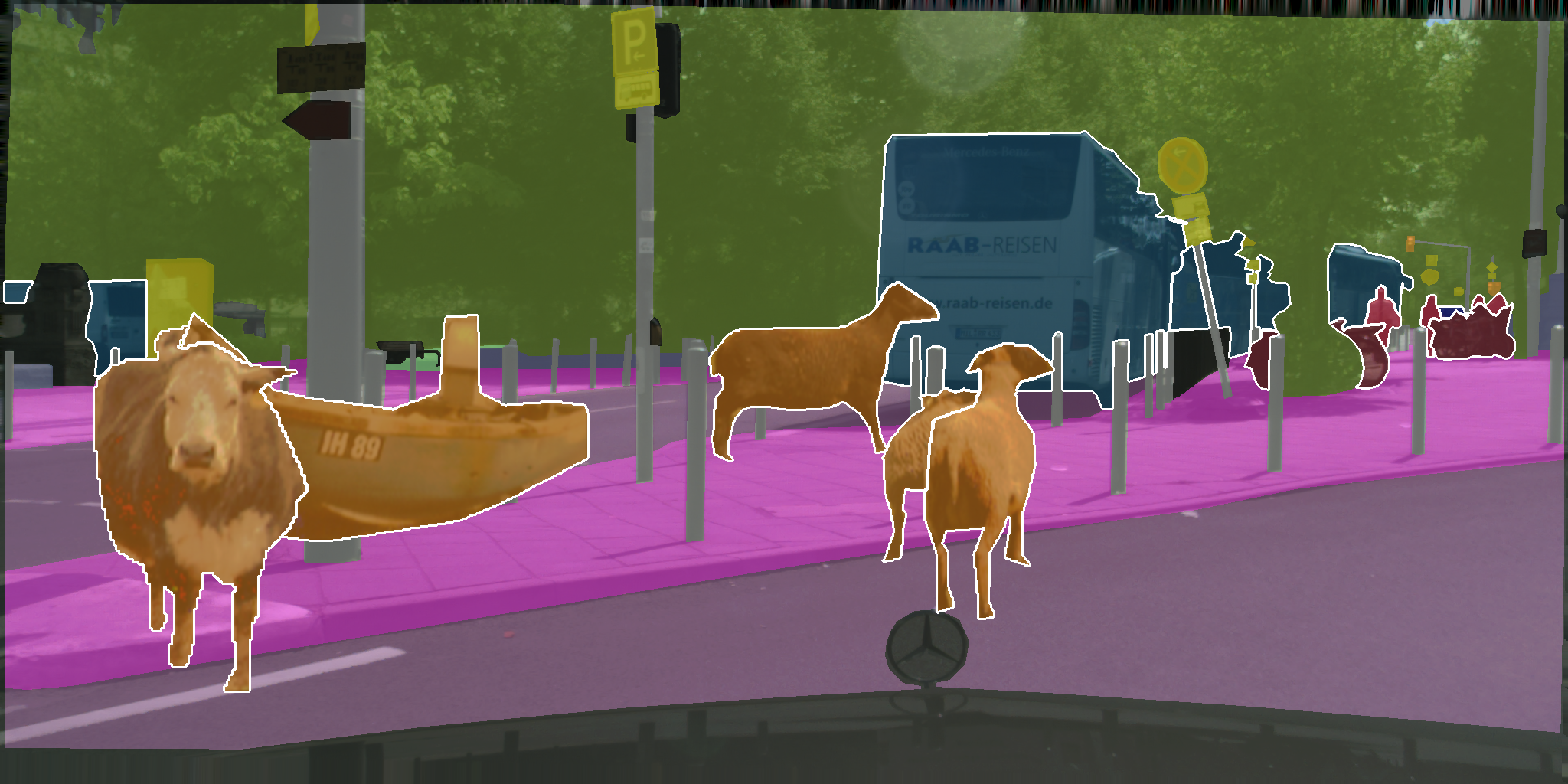}} & \raisebox{-0.4\height}{\includegraphics[width=\linewidth,frame]{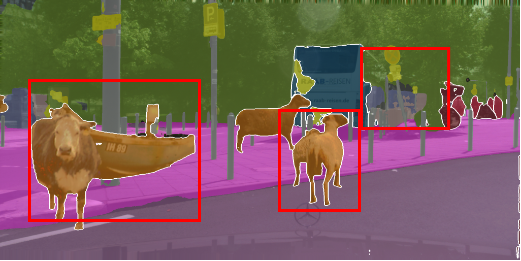}} &  \raisebox{-0.4\height}{\includegraphics[width=\linewidth]{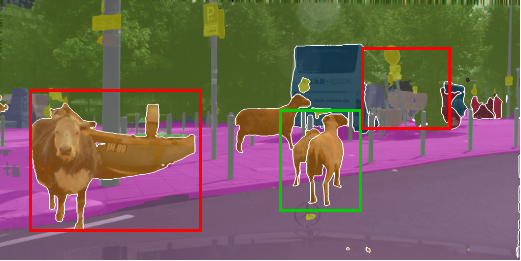}}\\
\\
\end{tabular}}
\caption{Qualitative panoptic out-of-distribution segmentation results of our proposed PoDS network in comparison to the state-of-the-art baseline DD-OPS~\cite{xu2022dual} on Cityscapes-OOD (a, d) and BDD100K-OOD (b, c) datasets.} 
\label{fig:visual_ablation}
\vspace{-0.3cm}
\end{figure*}



\begin{table}
\footnotesize 
\centering
\caption{Evaluation of various architectural components in \net. Results are presented on the Cityscapes-OOD test set. Subscripts $out$ and $in$ refer to out-of-distribution class and in-distribution classes. All scores are in [\%].}
\label{tab:ablarch}
\begin{tabular}{p{1.7cm}|p{1.0cm}p{0.6cm}p{0.6cm}}
\toprule
Model &  POD-Q  & PQ$_{out}$ & PQ$_{in}$\\
\midrule
M1 & $28.2$&$16.2$&$49.3$ \\
M2 & $26.9$&$15.3$&$47.5$ \\
M3 & $47.1$&$38.4$&$58.0$ \\
M4 (\net) & $\mathbf{53.4}$  & $\mathbf{45.9}$  & $\mathbf{62.2}$\\
\bottomrule
\end{tabular}
\vspace{-0.5cm}
\end{table}


\subsection{Ablation Study}

\subsubsection{Detailed Study on the PoDS Architecture}
\label{subsubsec:arch}
While developing the PoDS architecture, we incorporated various components to address specific challenges. \tabref{tab:ablarch} presents four model configurations, labeled M$_i$, to determine the impact of each component. The M1 configuration uses the base network PAPS$^*$ with OOD classes as an extra class, trained from scratch with data augmentation. We observe that M1 achieves a low POD-Q score of $28.2$ as it tries to cover both in-distribution and out-of-distribution classes, leading to poor generalization on the test set for unseen \textit{OOD} classes. In M2, we incorporate the dual predictive head architecture of PoDS into M1 and train it with our alignment-mismatch loss. However, this leads to a drop of $1.3$ in the POD-Q score compared to M1, indicating that the pretrained backbone does not encode OOD objects effectively enough for the new decoders and heads to learn. As a result, the alignment-mismatch loss hinders the performance of M2. In M3, we incorporate the OOD contextual module into M2. The notable performance improvement compared to M2 indicates that by learning highly discriminative features through the simplified task of OOD classification and segmentation, the dual predictive head, combined with the alignment-mismatch loss, prioritizes understanding what lies outside the in-distribution rather than trying to model the distribution of out-of-distribution objects. Finally, in M4, we incorporate the dynamic module into the non-pretrained decoders of M3. The results of M4, with a POD-Q score of $50.5$, underscore the benefits of learning features not previously encompassed in the in-distribution feature space. As elaborated in \secref{subsubsec:dynamic}, this is achieved by leveraging pretrained weight offsets and dynamically transitioning between the robust knowledge base of the pretrained decoder for in-distribution classes and the decoder trained to recognize features in the presence of both in-distribution and out-of-distribution objects. Moreover, the improvements from M$_{i-1}$ to M$_{i}$ result not only from the new additions but also from their synergy with the existing modules. M4 embodies our proposed PoDS architecture.

\begin{table}
\footnotesize 
\centering
\caption{Evaluation of the top two baselines with PoDS' data augmentation. Results are presented on the Cityscapes-OOD test set. Subscript $out$ and $in$ refers to out-of-distribution and in-distribution classes. All scores are in [\%].}
\label{tab:augeval}
\begin{tabular}{l|c|ccc}
\toprule
Model & Data Augmentation & POD-Q & PQ$_{out}$ &  PQ$_{in}$ \\
\midrule
Meta-OOD~\cite{chan2021entropy} &  &$41.7$ &   $31.3$ &$55.6$    \\
DD-OPS~\cite{xu2022dual}   &  & $46.1$ &   $36.1$ &$58.7$  \\
\midrule
Meta-OOD~\cite{chan2021entropy} & \checkmark &$43.0$ &   $32.7$ &$56.5$    \\
DD-OPS~\cite{xu2022dual}   &\checkmark &$48.5$ &   $39.4$ &$59.8$    \\
\bottomrule
\end{tabular}
\vspace{-0.3cm}
\end{table}


\subsubsection{Impact of Data Augmentation Strategy}
\label{subsubsec:aug}

We evaluate our proposed augmentation learning strategy, which introduces diverse out-of-distribution objects not sourced from a fixed distribution. We apply this strategy to the top-performing baselines, Meta-OOD and DD-OPS, on the Cityscapes-OOD dataset. The results in \tabref{tab:augeval} show that their POD-Q scores increase by $1.3$ for Meta-ODD and $2.4$ for DD-OPS post-augmentation without affecting the PQ$_{in}$ performance. This increase stems from the progressive exposure of augmented OOD data that allows the network to initially discern objects distinctly different from the in-distribution objects. However, comparing these improvements with M1 and M2's performance from \secref{subsubsec:arch}, we infer that while augmentation does contribute to the improved performance of PoDS, the other network components of PoDS are equally crucial for its significant improvement.

\begin{table}
\setlength\tabcolsep{1.2pt}
\footnotesize 
\centering
\caption{Evaluation of best panoptic out-of-distribution segmentation methods on the Cityscapes val set. Subscript $out$ and $in$ refers to out-of-distribution and in-distribution classes. Subscript $base$ refers to the base panoptic segmentation network. All scores are in [\%].}
\begin{subtable}{.53\linewidth}
\caption{Panoptic out-of-distribution performance using bicycle and motorcycle classes as OOD.}
\label{tab:real}
\begin{tabular}{l|ccc}
\toprule
Model & POD-Q & PQ$_{out}$ & PQ$_{in}$ \\
\midrule
Meta-OOD~\cite{chan2021entropy}  &$39.1$ & $27.1$ &$56.3$ \\
DD-OPS~\cite{xu2022dual}     & $44.7$ & $33.4$ &$59.8$ \\
PoDS   &$50.3$ & $39.8$ & $63.6$  \\
\bottomrule
\end{tabular}
\end{subtable}
\begin{subtable}{.35\linewidth}
\caption{Influence of OOD seg. on in-distribution performance.}
\label{tab:augeval_}
\begin{tabular}{l|cc}
\toprule
Model & PQ & PQ${_{base}}$  \\
\midrule
Meta-OOD~\cite{chan2021entropy}  &$60.7$ & $63.9$ \\
DD-OPS~\cite{xu2022dual}     & $62.5$ & $63.9$ \\
PoDS   &$63.1$ & $63.7$ \\
\bottomrule
\end{tabular}
\end{subtable}
\end{table}

\subsubsection{Evaluation in Real-World OOD Scenarios}
\label{subsubsec:real}

In this experiment, we evaluate the utility of our models and baselines in real-world settings using the Cityscapes dataset. We include two \textit{thing} classes, bicycle, and motorcycle, from the eight Cityscapes \textit{thing} classes as part of the OOD class. We exclude any image from the training set containing at least one instance of this OOD class, reducing the training set from 2975 to 2620 images. This exclusion ensures that the bicycle and motorcycle classes are treated as unseen OOD objects during evaluation. The results, presented in \tabref{tab:real} demonstrate that PoDS consistently outperforms the top two baselines by a substantial margin, reinforcing the findings from \tabref{tab:cityOODEvaluation} and underscoring its applicability to real-world OOD scenarios. In \secref{sec:sreal}, we further qualitatively demonstrate the generalization ability of PoDS in real-world driving scenarios using our in-house data collected in Freiburg.

\subsubsection{Influence of OOD Segmentation on In-Distribution Performance}
\label{subsubsec:panop}

\begin{table}
\centering
\caption{Performance of PoDS models trained on Cityscapes but evaluated on BDD100K val set and BDD100K-OOD test set. All scores are in [\%].}
\label{tab:cross_eval}
\footnotesize
\begin{tabular}{l|c|c|ccc}
\toprule
Training & Method & \multicolumn{4}{c}{Evaluation Dataset} \\
\cmidrule{3-6}
Dataset &  &\multicolumn{1}{c|}{BDD100K} & \multicolumn{3}{c}{BDD100K-OOD} \\
\cmidrule{3-6}
&  & PQ & POD-Q & PQ$_{in}$ & PQ$_{out}$  \\
\midrule
\multirow{2}{*}{Cityscapes} & PAPS$^*$ & $39.6$  & $-$ & $-$ & $-$  \\
                            & PoDS & $38.9$ & $28.1$ & $38.2$ & $20.6$ \\
\bottomrule
\end{tabular}
\vspace{-0.3cm}
\end{table}

We first study the impact of learning panoptic out-of-distribution segmentation on network performance when only in-distribution classes are present in the input. We compare with the top three methods: Meta-OOD, DD-OPS, and PoDS, and also report the performance of their base panoptic segmentation networks. From the results shown in~\tabref{tab:augeval_}, we observe that the PQ score of Meta-OOD substantially decreases, while DD-OPS and PoDS show a smaller drop in performance. However, PoDS shows the least drop of $0.6$, demonstrating the ability to segment out-of-distribution objects while preserving the in-distribution class knowledge.
Subsequently, we evaluate the generalization ability of PoDS by training it on the Cityscapes dataset and evaluating it on the BDD100K dataset. Results from this experiment presented in \tabref{tab:cross_eval} show that PoDS performs nearly as well as its base network PAPS$^*$, achieving a POD-Q score of $28.1$ on BDD100K-OOD and a PQ of $38.9$ on BDD100K. This highlights the ability of PoDS to infer known semantic class boundaries from Cityscapes, constrained only by its base network's performance. We anticipate further advancements in this field by the robotics community will surpass these limitations in the future.



\subsection{Qualitative Evaluations}
\label{subsubsec:qualitative}

We qualitatively compare the performance of PoDS with the best-performing baseline DD-OPS~\cite{xu2022dual} as illustrated in \figref{fig:visual_ablation}. We observe that DD-OPS misclassifies OOD objects with known semantic classes, while PoDS excels at distinguishing them. PoDS exploits its dynamic module and the alignment-mismatch strategy to identify OOD features based on known semantic characteristics, enabling it to accurately distinguish between OOD objects, and bicycles and cars. However, we observe that PoDS struggles with cluttered OOD objects and is limited by its base network, as shown in \figref{fig:visual_ablation} (d), misclassifying a bus as \textit{stuff} due to its size and sample constraints. We hope that this work encourages solutions in the future to address these limitations.

\section{Conclusion}

In this work, we introduced the panoptic out-of-distribution segmentation task, proposed two suitable datasets, established an interpretable evaluation metric, and adapted several open-set and semantic out-of-distribution segmentation methods for baselines. We also proposed the novel PoDS architecture, which sets a new benchmark in performance.  It also demonstrates the feasibility of incorporating OOD segmentation without a significant drop in in-distribution performance. We presented an extended evaluation of each module that we used in our network with quantitative and qualitative evaluations that demonstrate their utility. Our novel framework shows the feasibility of this crucial and holistic scene parsing task and we aim that our publicly released datasets and benchmark facilitate further research in this direction.

\footnotesize
\bibliographystyle{IEEEtran}
\bibliography{references}

\begin{thebibliography}{10}
\providecommand{\url}[1]{#1}
\csname url@rmstyle\endcsname
\providecommand{\newblock}{\relax}
\providecommand{\bibinfo}[2]{#2}
\providecommand\BIBentrySTDinterwordspacing{\spaceskip=0pt\relax}
\providecommand\BIBentryALTinterwordstretchfactor{4}
\providecommand\BIBentryALTinterwordspacing{\spaceskip=\fontdimen2\font plus
\BIBentryALTinterwordstretchfactor\fontdimen3\font minus \fontdimen4\font\relax}
\providecommand\BIBforeignlanguage[2]{{%
\expandafter\ifx\csname l@#1\endcsname\relax
\typeout{** WARNING: IEEEtran.bst: No hyphenation pattern has been}%
\typeout{** loaded for the language `#1'. Using the pattern for}%
\typeout{** the default language instead.}%
\else
\language=\csname l@#1\endcsname
\fi
#2}}

\bibitem{vodisch2022continual}
N.~V{\"o}disch, D.~Cattaneo, W.~Burgard, and A.~Valada, ``Continual slam: Beyond lifelong simultaneous localization and mapping through continual learning,'' in \emph{The Int. Symposium of Robotics Research}, 2022, pp. 19--35.

\bibitem{gosala2023skyeye}
N.~Gosala, K.~Petek, P.~L. Drews-Jr, W.~Burgard, and A.~Valada, ``Skyeye: Self-supervised bird's-eye-view semantic mapping using monocular frontal view images,'' in \emph{Proc.~of the IEEE Conf.~on Computer Vision and Pattern Recognition}, 2023, pp. 14\,901--14\,910.

\bibitem{kirillov2019panoptic}
A.~Kirillov, K.~He, R.~Girshick, C.~Rother, and P.~Doll{\'a}r, ``Panoptic segmentation,'' in \emph{Proc.~of the IEEE Conf.~on Computer Vision and Pattern Recognition}, 2019, pp. 9404--9413.

\bibitem{bozhinoski2019safety}
D.~Bozhinoski, D.~Di~Ruscio, I.~Malavolta, P.~Pelliccione, and I.~Crnkovic, ``Safety for mobile robotic systems: A systematic mapping study from a software engineering perspective,'' \emph{Journal of Systems and Software}, vol. 151, pp. 150--179, 2019.

\bibitem{hwang2021exemplar}
J.~Hwang, S.~W. Oh, J.-Y. Lee, and B.~Han, ``Exemplar-based open-set panoptic segmentation network,'' in \emph{Proc.~of the IEEE Conf.~on Computer Vision and Pattern Recognition}, 2021, pp. 1175--1184.

\bibitem{mohan2020efficientps}
R.~Mohan and A.~Valada, ``Efficientps: Efficient panoptic segmentation,'' \emph{Int.~Journal of Computer Vision}, vol. 129, no.~5, pp. 1551--1579, 2021.

\bibitem{kirillov2019bpanoptic}
A.~Kirillov, R.~Girshick, K.~He, and P.~Doll{\'a}r, ``Panoptic feature pyramid networks,'' in \emph{Proc.~of the IEEE Conf.~on Computer Vision and Pattern Recognition}, 2019, pp. 6399--6408.

\bibitem{cheng2020panoptic}
B.~Cheng, M.~D. Collins, Y.~Zhu, T.~Liu, T.~S. Huang, H.~Adam, and L.-C. Chen, ``Panoptic-deeplab: A simple, strong, and fast baseline for bottom-up panoptic segmentation,'' in \emph{Proc.~of the IEEE Conf.~on Computer Vision and Pattern Recognition}, 2020.

\bibitem{uhrig2018box2pix}
J.~Uhrig, E.~Rehder, B.~Fr{\"o}hlich, U.~Franke, and T.~Brox, ``Box2pix: Single-shot instance segmentation by assigning pixels to object boxes,'' in \emph{IEEE Intelligent Vehicles Symposium}, 2018, pp. 292--299.

\bibitem{mohan22perceiving}
R.~Mohan and A.~Valada, ``Perceiving the invisible: Proposal-free amodal panoptic segmentation,'' \emph{IEEE Robotics and Automation Letters}, vol.~7, no.~4, pp. 9302--9309, 2022.

\bibitem{mohan2022amodal}
------, ``Amodal panoptic segmentation,'' in \emph{Proc.~of the IEEE Conf.~on Computer Vision and Pattern Recognition}, 2022, pp. 21\,023--21\,032.

\bibitem{hendrycks2016baseline}
D.~Hendrycks and K.~Gimpel, ``A baseline for detecting misclassified and out-of-distribution examples in neural networks,'' \emph{arXiv preprint arXiv:1610.02136}, 2016.

\bibitem{hendrycks2019scaling}
D.~Hendrycks, S.~Basart, M.~Mazeika, M.~Mostajabi, J.~Steinhardt, and D.~Song, ``Scaling out-of-distribution detection for real-world settings,'' \emph{arXiv preprint arXiv:1911.11132}, 2019.

\bibitem{kendall2017uncertainties}
A.~Kendall and Y.~Gal, ``What uncertainties do we need in bayesian deep learning for computer vision?'' \emph{Advances in Neural Information Processing Systems}, vol.~30, 2017.

\bibitem{gal2016dropout}
Y.~Gal and Z.~Ghahramani, ``Dropout as a bayesian approximation: Representing model uncertainty in deep learning,'' in \emph{Int.~Conf.~on Machine Learning}, 2016, pp. 1050--1059.

\bibitem{lakshminarayanan2017simple}
B.~Lakshminarayanan, A.~Pritzel, and C.~Blundell, ``Simple and scalable predictive uncertainty estimation using deep ensembles,'' \emph{Advances in Neural Information Processing Systems}, vol.~30, 2017.

\bibitem{blum2019fishyscapes}
H.~Blum, P.-E. Sarlin, J.~Nieto, R.~Siegwart, and C.~Cadena, ``Fishyscapes: A benchmark for safe semantic segmentation in autonomous driving,'' in \emph{Proc. of the IEEE/CVF Int. Conf. on Computer Vision Workshops}, 2019.

\bibitem{chan2021entropy}
R.~Chan, M.~Rottmann, and H.~Gottschalk, ``Entropy maximization and meta classification for out-of-distribution detection in semantic segmentation,'' in \emph{Proc.~of the IEEE Conf.~on Computer Vision and Pattern Recognition}, 2021, pp. 5128--5137.

\bibitem{xia2020synthesize}
Y.~Xia, Y.~Zhang, F.~Liu, W.~Shen, and A.~L. Yuille, ``Synthesize then compare: Detecting failures and anomalies for semantic segmentation,'' in \emph{Europ.~Conf.~on Computer Vision}, 2020, pp. 145--161.

\bibitem{liang2017enhancing}
S.~Liang, Y.~Li, and R.~Srikant, ``Enhancing the reliability of out-of-distribution image detection in neural networks,'' \emph{arXiv preprint arXiv:1706.02690}, 2017.

\bibitem{besnier2021triggering}
V.~Besnier, A.~Bursuc, D.~Picard, and A.~Briot, ``Triggering failures: Out-of-distribution detection by learning from local adversarial attacks in semantic segmentation,'' in \emph{Int.~Conf.~on Computer Vision}, 2021, pp. 15\,701--15\,710.

\bibitem{xu2022dual}
H.-M. Xu, H.~Chen, L.~Liu, and Y.~Yin, ``Dual decision improves open-set panoptic segmentation,'' in \emph{British Mac.~Vision Conf.}, 2022.

\bibitem{gupta2019lvis}
A.~Gupta, P.~Dollar, and R.~Girshick, ``Lvis: A dataset for large vocabulary instance segmentation,'' in \emph{Proc.~of the IEEE Conf.~on Computer Vision and Pattern Recognition}, 2019, pp. 5356--5364.

\bibitem{cordts2016cityscapes}
M.~Cordts, M.~Omran, S.~Ramos, T.~Rehfeld, M.~Enzweiler, R.~Benenson, U.~Franke, S.~Roth, and B.~Schiele, ``The cityscapes dataset for semantic urban scene understanding,'' in \emph{Proc.~of the IEEE Conf.~on Computer Vision and Pattern Recognition}, 2016, pp. 3213--3223.

\bibitem{yu2020bdd100k}
F.~Yu, H.~Chen, X.~Wang, W.~Xian, Y.~Chen, F.~Liu, V.~Madhavan, and T.~Darrell, ``Bdd100k: A diverse driving dataset for heterogeneous multitask learning,'' in \emph{Proc.~of the IEEE Conf.~on Computer Vision and Pattern Recognition}, 2020, pp. 2636--2645.

\bibitem{xu2022regnet}
J.~Xu, Y.~Pan, X.~Pan, S.~Hoi, Z.~Yi, and Z.~Xu, ``Regnet: self-regulated network for image classification,'' \emph{IEEE Transactions on Neural Networks and Learning Systems}, 2022.

\bibitem{wang2022freesolo}
X.~Wang, Z.~Yu, S.~De~Mello, J.~Kautz, A.~Anandkumar, C.~Shen, and J.~M. Alvarez, ``Freesolo: Learning to segment objects without annotations,'' in \emph{Proc.~of the IEEE Conf.~on Computer Vision and Pattern Recognition}, 2022, pp. 14\,176--14\,186.

\end{thebibliography}

\clearpage
\renewcommand{\baselinestretch}{1}
\setlength{\belowcaptionskip}{0pt}

\begin{strip}
\begin{center}
\vspace{-5ex}
\textbf{\LARGE \bf
Panoptic Out-of-Distribution Segmentation} \\
\vspace{3ex}

\Large{\bf- Supplementary Material -}\\
\vspace{0.4cm}
\normalsize{Rohit Mohan, 
        Kiran Kumaraswamy, 
        Juana Valeria Hurtado, 	
        Kürsat Petek, 
        and~Abhinav Valada}
\end{center}
\end{strip}

\setcounter{section}{0}
\setcounter{equation}{0}
\setcounter{figure}{0}
\setcounter{table}{0}
\setcounter{page}{1}
\makeatletter

\renewcommand{\thesection}{S.\arabic{section}}
\renewcommand{\thesubsection}{S.\arabic{subsection}}
\renewcommand{\thetable}{S.\arabic{table}}
\renewcommand{\thefigure}{S.\arabic{figure}}


\let\thefootnote\relax\footnote{Department of Computer Science, University of Freiburg, Germany.\\
Project page: \url{http://pods.cs.uni-freiburg.de}
}%
\normalsize
In this supplementary material, we begin with a detailed description of the POD-Q metric in \secref{sec:smetric}, followed by additional details on our PoDS architecture in \secref{sec:sarch}. Finally, in \secref{sec:sreal}, we provide further evidence supporting the pertinence of our proposed panoptic out-of-distribution (PoDs) task in real-world scenarios. 

\section{Panoptic Out-of-Distribution Quality}
\label{sec:smetric}

The objective of panoptic out-of-distribution segmentation is to accurately classify and segment both in-distribution and out-of-distribution objects in a scene. To effectively evaluate the performance of this task, we need to equally assess the performance of both object categories while having the ability to distinguish the category-specific results. Furthermore, an ideal evaluation metric should be interpretable and easy to implement, promoting transparency and simplicity. In this direction, we propose the Panoptic Out-of-Distribution Quality (POD-Q), a metric based on the popular panoptic quality (PQ) metric~\cite{kirillov2019panoptic}.
To compute the POD-Q metric, we first compute PQ$_{OOD}$ as the PQ for the $OOD$ classes $o \in O$ given the predicted object segments $P$ and their ground truth object segments $G$ as follows:
\begin{equation}
PQ_{OOD} = \frac{\sum_{(p,g) \in TP_{o}, } IoU(p,g)} {|TP_o|+\frac{1}{2}|FP_o|+\frac{1}{2}|FN_o|},
\end{equation}

Subsequently, we compute the PQ score for all the in-distribution semantic classes PQ$_{in}$ as
\begin{equation}
PQ_{in} = \frac{1}{|C|}\sum_{c \in C}\frac{\sum_{(p,g) \in TP_{c}, } IoU(p,g)} {|TP_c|+\frac{1}{2}|FP_c|+\frac{1}{2}|FN_c|},
\end{equation}
where $C$ is the set of in-distribution semantic classes. 
For both PQ$_{OOD}$ and PQ$_{in}$, the true positives ($TP_{i}$), false positives ($FP_i$), and false negatives ($FN_i$) are defined as
\begin{align}
TP_i &= \{p_i \in \nset{P} \;|\; IoU(p_i, g_i) > 0.5, \; \forall \;\; g_c \in \nset{G}\}, \\
FP_i &= \{p_i \in \nset{P} \;|\; IoU(p_i, g) <= 0.5, \; \forall \; g \in \nset{G}\}, \\
FN_i &= \{g_i \in \nset{G} \;|\; IoU(g_i, p) <= 0.5, \; \forall \; p \in \nset{P}\}.
\end{align}
where $i$ takes the value of $o$ for OOD class and $c$ for  in-distribution semantic classes with $c \in C$.
Finally, we compute POD-Q as the geometric mean between PQ$_{OOD}$ and PQ$_{in}$, as
\begin{equation}
POD\text{-}Q = (PQ_{OOD} \times PQ_{in})^{\frac{1}{2}}.
\end{equation}

\begin{figure*}
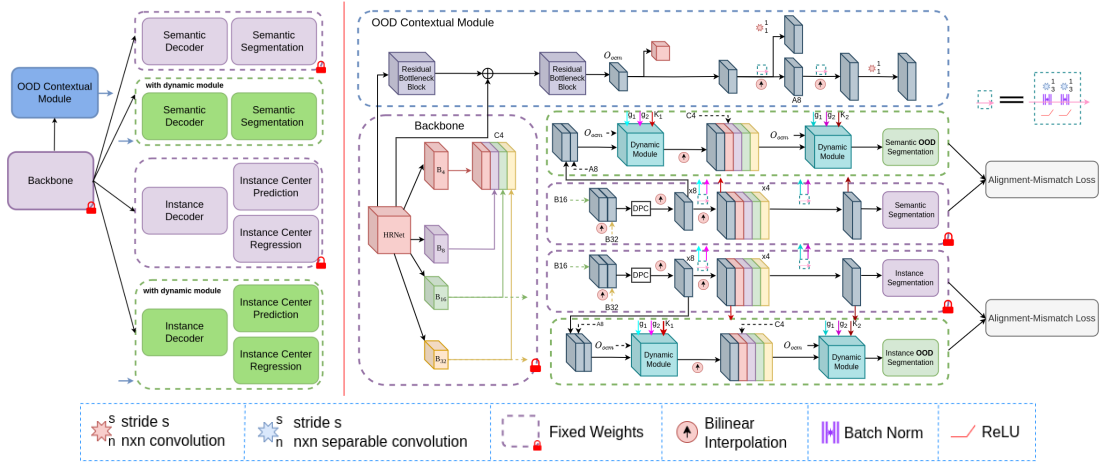

    \centering
    \begin{subfigure}[b]{\linewidth}
        \centering
        \includegraphics[width=0.8\linewidth]{figures/test_new_model5.png}
        \label{APSNEt_arch_supp}
    \end{subfigure}
        \hfill
    \begin{subfigure}[b]{0.7\linewidth}
       \centering
        \includegraphics[width=\textwidth]{figures/legend.png}
    \end{subfigure}
    \caption{Illustration of our proposed PoDS architecture that consists of a shared backbone with an OOD contextual module and symmetrical task-specific decoder arranged in a dual configuration setup to facilitate an alignment-mismatch learning strategy. The shared backbone learns robust feature representations for in-distribution semantic categories while the OOD contextual module supports both global and local features for OOD objects. The network comprises symmetrical semantic and instance decoders that include dynamic modules to adaptively balance the features between in- and out-distribution representations.}
    \label{fig:network_supp}
\end{figure*}

\section{PoDS Architecture}
\label{sec:sarch}

This section describes the base network in more detail and the OOD contextual module of the PoDS architecture proposed in \figref{fig:network_supp}.

\subsection{Base Network}
The PAPS$^*$ architecture which is the base network of PoDS, employs HRNet as its backbone which is designed to retain high-resolution information throughout the network. The backbone generates four parallel feature map outputs at scales $\times4$, $\times8$, $\times16$, and $\times32$ with respect to the input, named B\textsubscript{4}, B\textsubscript{8}, B\textsubscript{16}, and B\textsubscript{32}. These feature maps are further upsampled to a resolution of $\times4$ and combined to create the C\textsubscript{4} feature maps. The outputs from the backbone are fed to both the semantic and instance decoders, which have a similar architecture. The decoders take B\textsubscript{32}, B\textsubscript{16}, and C\textsubscript{4} as inputs. The B\textsubscript{32} feature maps are first upsampled to $\times16$ resolution and concatenated with B\textsubscript{16}. This result is then fed into the dense prediction cell (DPC). The output from the DPC is then further upsampled to $\times8$ resolution and processed through two consecutive $3\times3$ depthwise-separable convolutions. In the next step, we upsample the output ($\times4$) and concatenate it with C4. This is then followed by two sequential $3\times3$ depth-wise separable convolutions before being fed to the task-specific heads. Following, each task-specific head has a similar design and consists of sequential layers of two $3\times3$ depthwise-separable convolutions, followed by a task-specific $1\times1$ predictor. In PoDS, the PAPS$^*$ part of the network is pretrained on the in-distribution panoptic segmentation dataset and its weights remain frozen during the entire panoptic out-of-distribution segmentation training.

\subsection{OOD Contextual Module}
The architecture of the OOD Contextual Module (OCM) is shown in {\figref{fig:network_supp}}. It consists of two residual bottleneck blocks comprising repeating units that employ group convolutions similar to the fourth and fifth stages in Regnet and a decoder. The output from the last layer of stage 2 of the backbone is processed by the first block of OCM, then concatenated with the output from the last layer of stage 3 and passed to the second block. Subsequently, the output of the second block (O\textsubscript{ocm}) is fed to a global average pooling layer which is followed by a $1\times1$ convolution layer that acts as a classification head. Following, the decoder takes O\textsubscript{ocm} as input and upsamples it by $\times 2$ scale followed by two sequential $3\times3$ depthwise-separable convolutions and an additional $\times 2$ scale upsampling. Subsequently, we process the resulting features {$A_8$} in two branches. The first branch uses a $1\times1$ convolution layer as an auxiliary semantic segmentation head. The second branch processes the output with two sequential $3\times3$ depthwise-separable convolutions followed by $\times2$ scale upsampling and $1\times1$ convolution layer as a second auxiliary semantic segmentation head for the upsampled features.

\section{Generalization in Real-World Scenarios}
\label{sec:sreal}

In this section, we aim to demonstrate the challenges associated with panoptic out-of-distribution segmentation and demonstrate the ability of our proposed network to segment OOD objects, as well as panoptic segmentation of in-distribution classes. To do so, we collect sequences of RGB images comprising real-world scenes captured in driving scenarios. We use the recorded data to qualitatively compare the performance of a conventional panoptic segmentation network and the proposed PoDs architecture. This comparison provides valuable insights into the differences between the two networks in scenarios that involve the segmentation of out-of-distribution objects. The results of this comparison show the ability of our proposed network to reason about objects that are very different from those presented during training. As a result, when having real-world images as an input, our network can provide a segmentation mask for unknown objects where the panoptic segmentation network incorrectly classifies them as background.

\subsection{Evaluation in Real-World Scenarios}

Panoptic segmentation is crucial for robot perception, as it enables robots to comprehend visual scenes they encounter by semantically segmenting and distinguishing instances from each other. Despite the holistic scene understanding provided by panoptic segmentation, it still presents limitations when identifying and segmenting out-of-distribution objects and maintaining the panoptic segmentation quality simultaneously. Panoptic out-of-distribution segmentation aims to address the gap between the current models trained for in-distribution tasks and the demands of robot perception in real-world scenarios. This task enables robots to recognize objects not seen during training, facilitating their operation in dynamic and unstructured environments with greater flexibility. Furthermore, this task also allows robots to preserve the panoptic segmentation quality. Panoptic out-of-distribution segmentation represents a significant step forward in making robots more suitable for real-world scenarios, thus improving their capability to comprehend visual scenes after deployment.

\subsection{Data Description}
In order to demonstrate the significance of panoptic out-of-distribution segmentation in practical settings, we conducted an additional collection of RGB data using a vehicle equipped with a sensor array using a FLIR Blackfly 23S3C camera with a resolution of $1920\times800$ pixels. The collected data consists of video sequences, as opposed to isolated images found in the Cityscapes-OOD and BDD100K-OOD datasets. Similar to these datasets, the collected data depicts driving scenarios relevant to autonomous driving applications. The driving scenes are composed of in-distribution and out-out-distribution objects. For in-distribution, the scene contains common objects such as cars, buildings, and vegetation. Additionally, for out-of-distribution, we placed uncommon objects, such as a fan, teddy bear, chair, kettle, suitcase, bottle, and trash bin on the road and sidewalk. 

\begin{figure*}
\centering
\footnotesize
{\renewcommand{\arraystretch}{0.5}
\begin{tabular}{P{0.4cm}P{5.5cm}P{5.5cm}P{5.5cm}}
&  \raisebox{-0.4\height}{Image} &  
\raisebox{-0.4\height}{PAPS~\cite{mohan22perceiving}} & \raisebox{-0.4\height}{PoDS (Ours)}\\
\\
\rot{(a)} 
& \raisebox{-0.4\height}{\includegraphics[width=\linewidth]{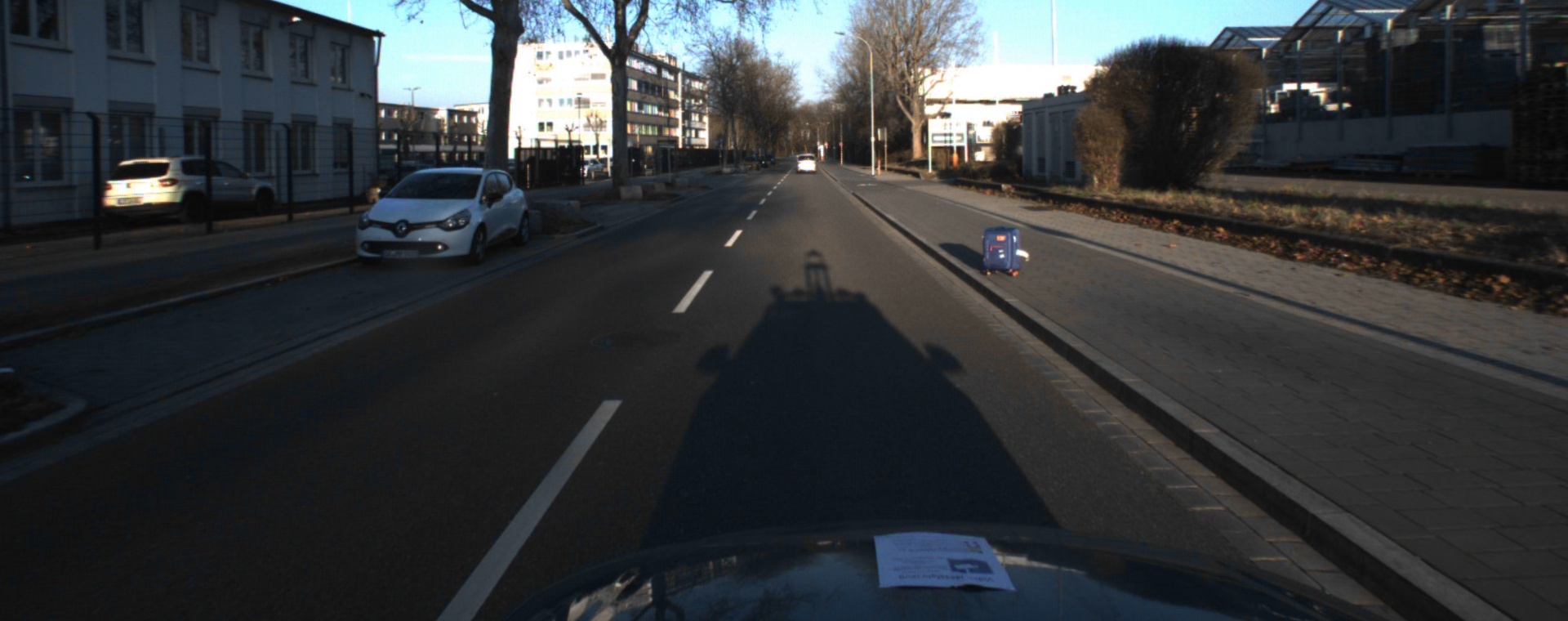}} & \raisebox{-0.4\height}{\includegraphics[width=\linewidth]{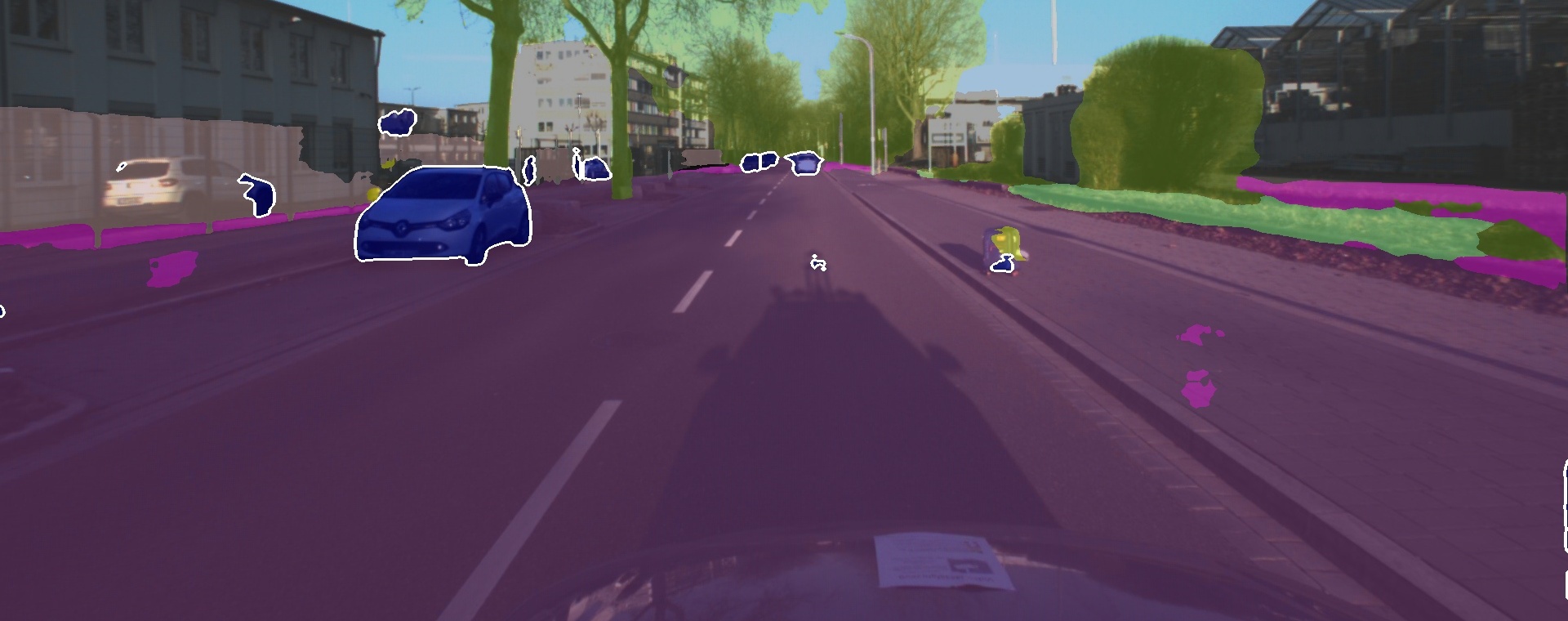}} & \raisebox{-0.4\height}{\includegraphics[width=\linewidth,frame]{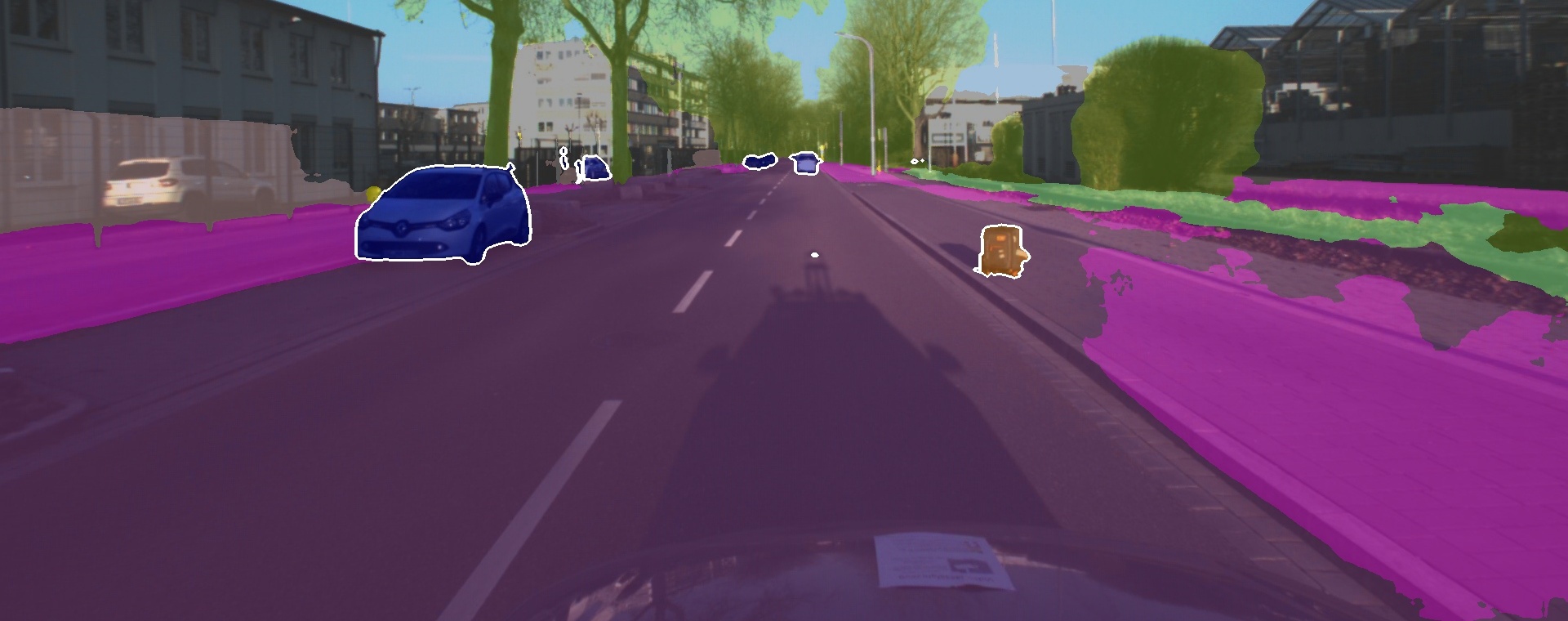}} 
\\
\\
\rot{(b)} 
& \raisebox{-0.4\height}{\includegraphics[width=\linewidth]{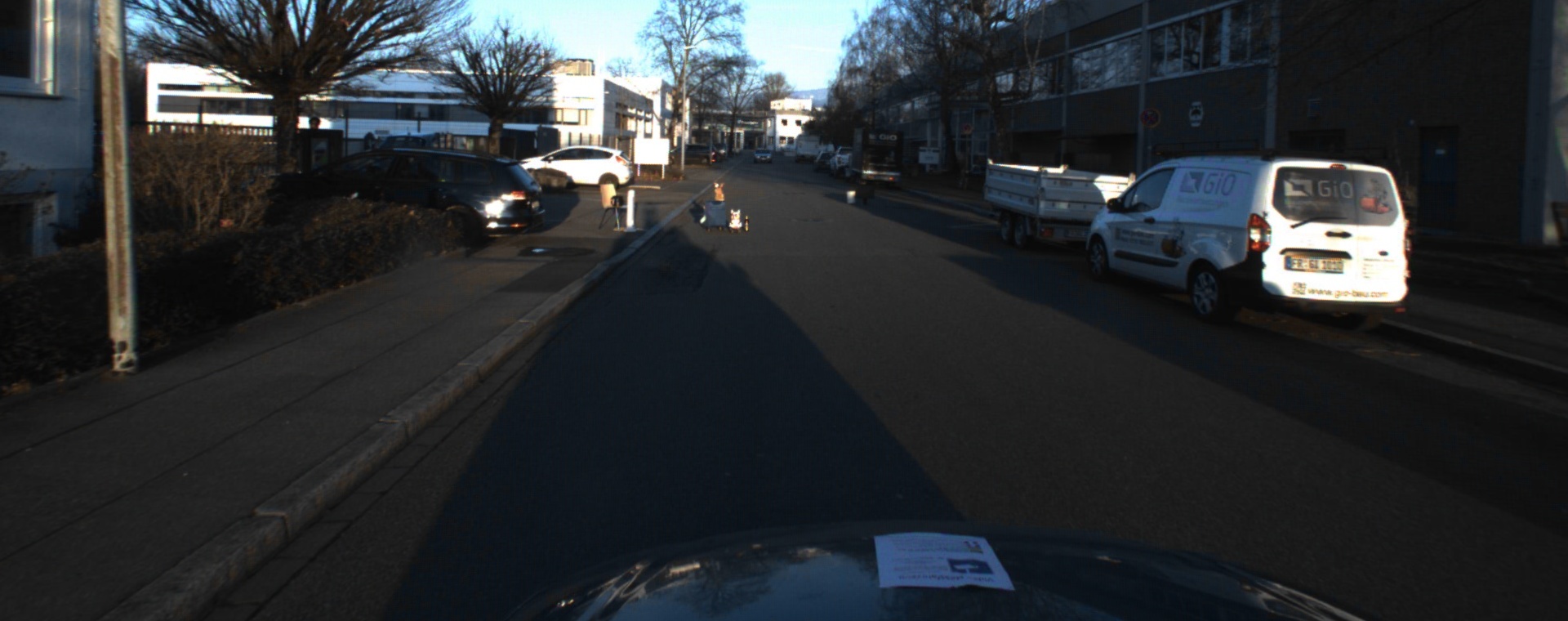}} & \raisebox{-0.4\height}{\includegraphics[width=\linewidth]{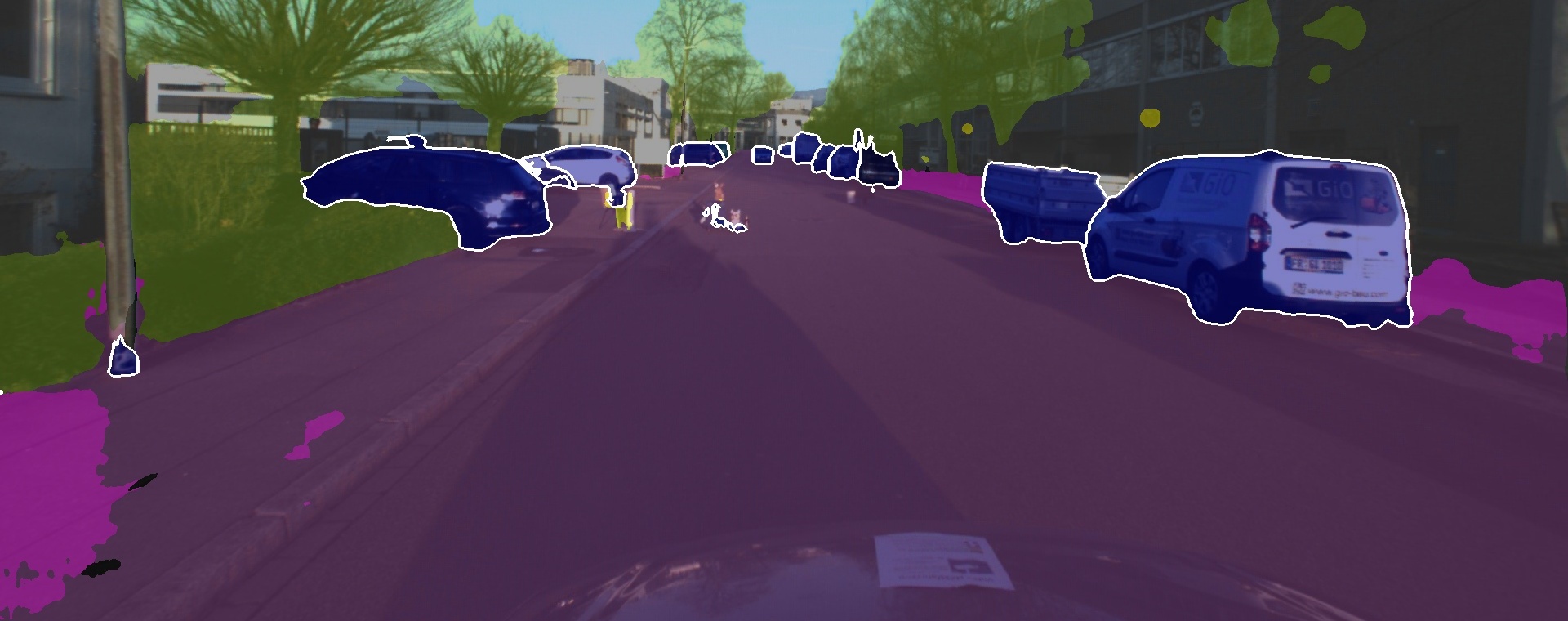}} & \raisebox{-0.4\height}{\includegraphics[width=\linewidth,frame]{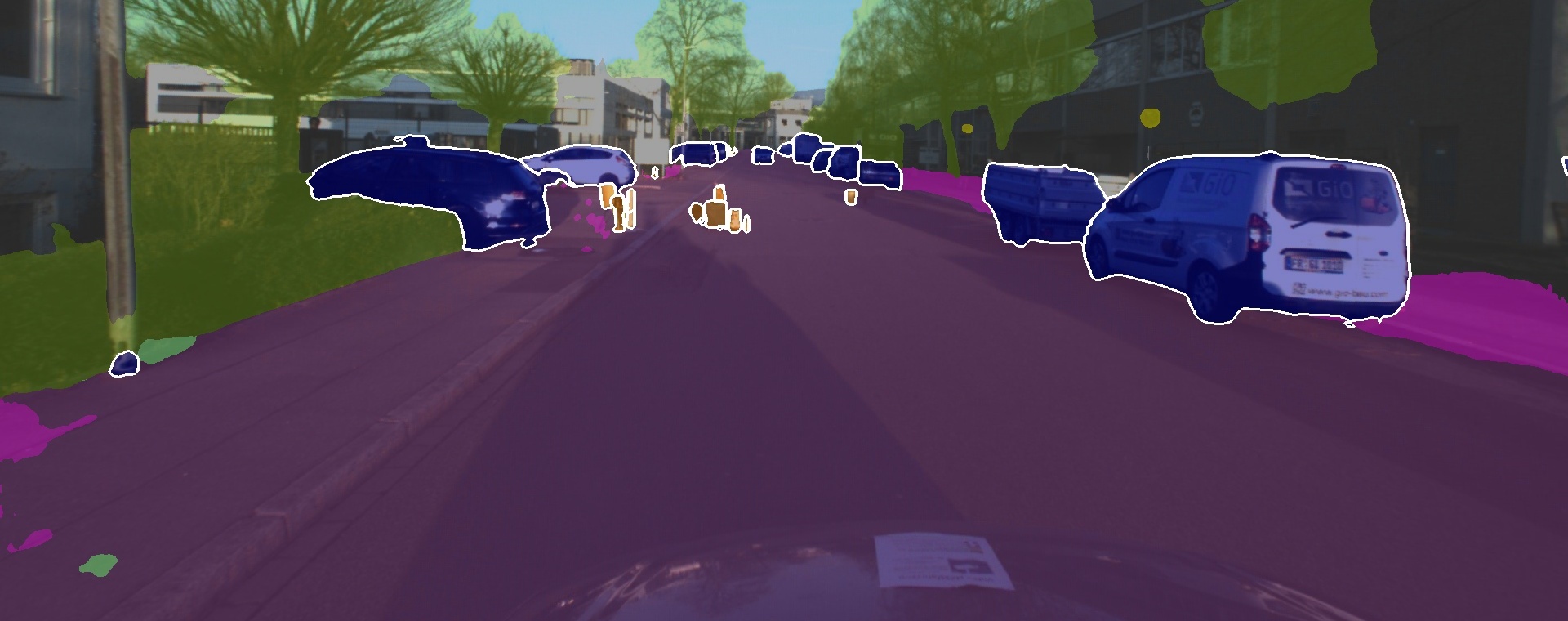}} 
\\
\\
\rot{(c)} 
& \raisebox{-0.4\height}{\includegraphics[width=\linewidth]{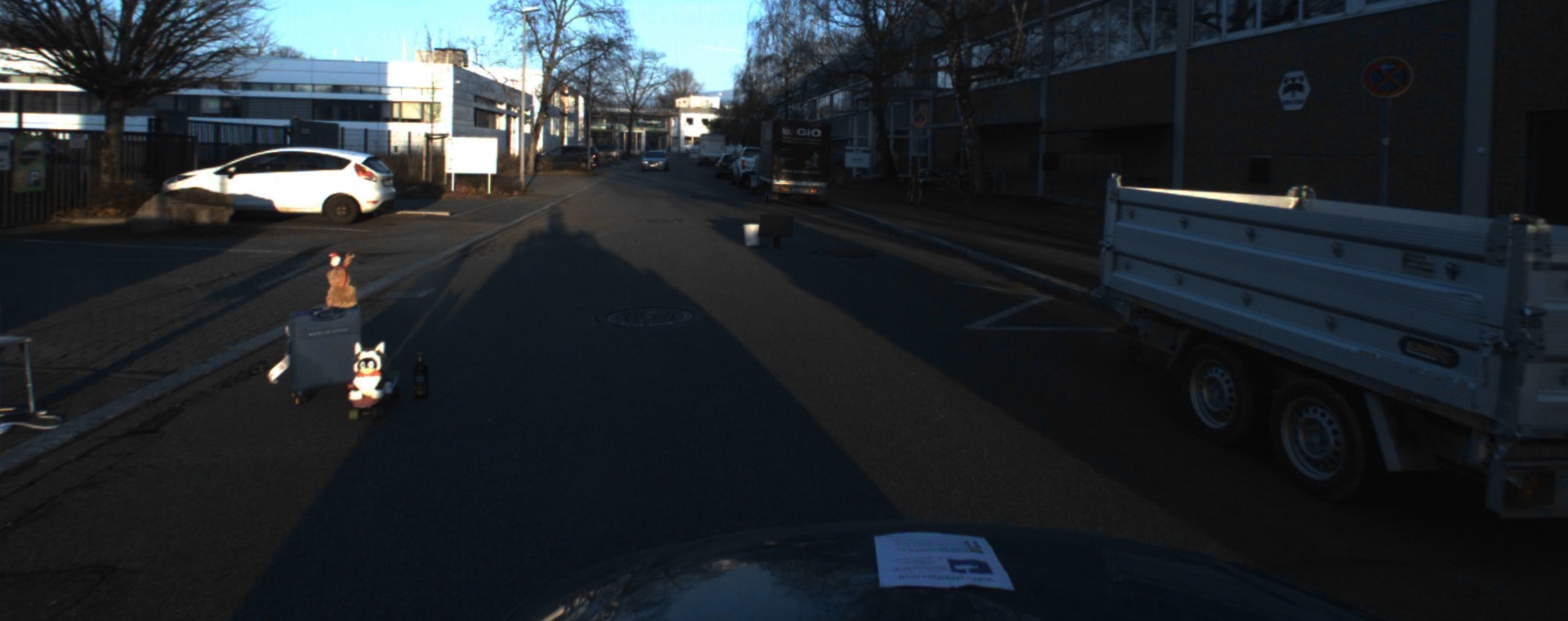}} & \raisebox{-0.4\height}{\includegraphics[width=\linewidth]{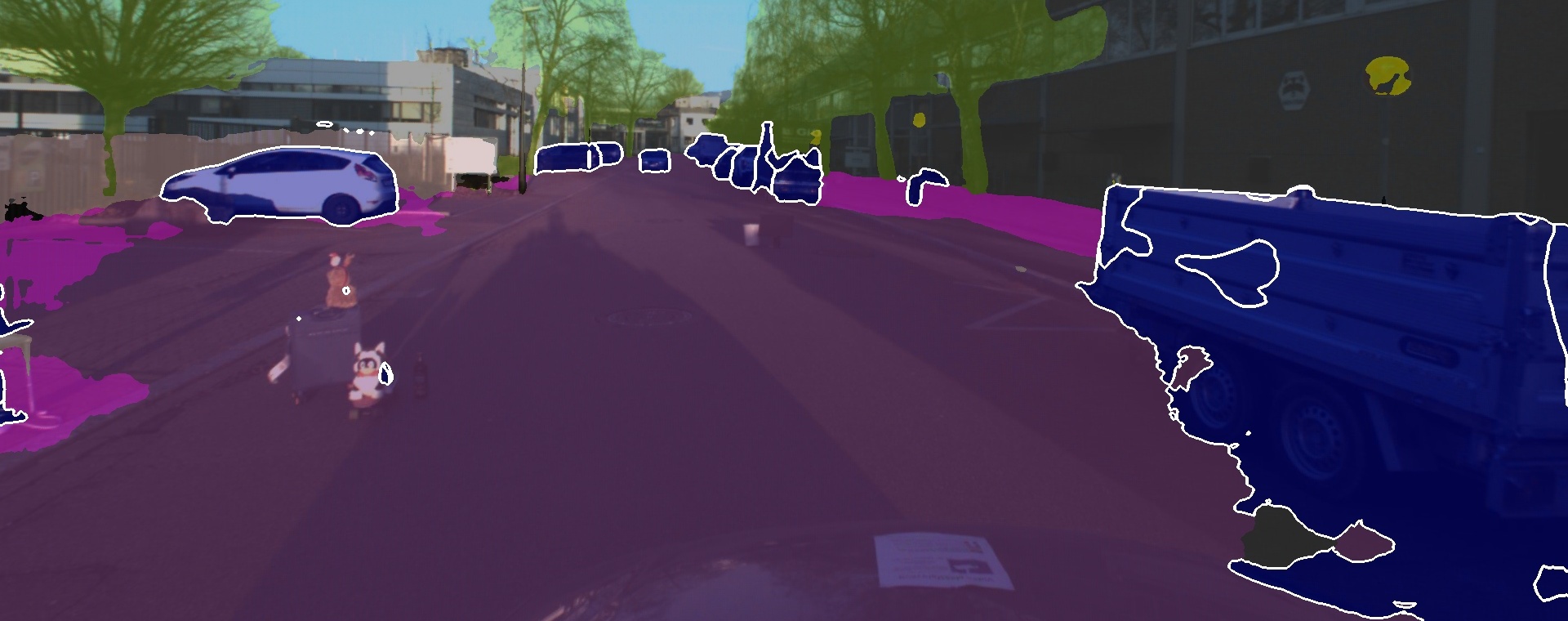}} & \raisebox{-0.4\height}{\includegraphics[width=\linewidth,frame]{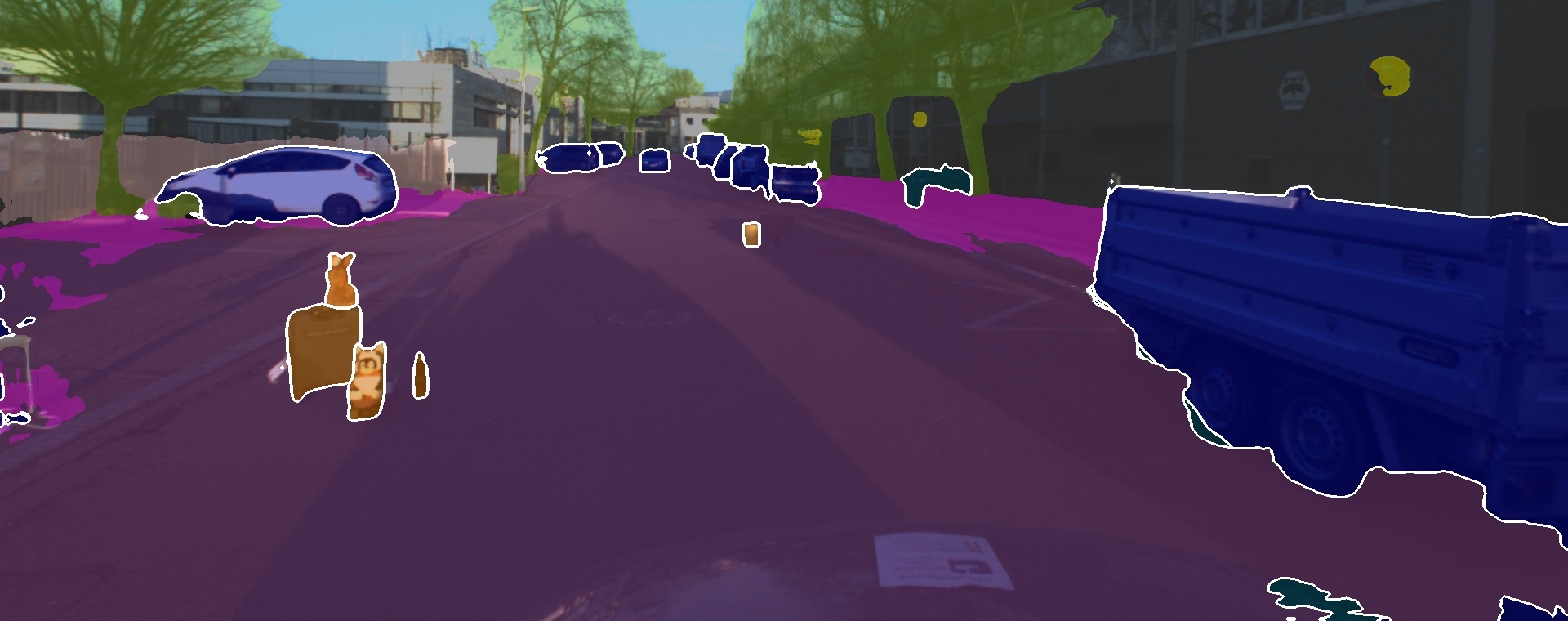}} 
\\
\\
\rot{(d)} 
& \raisebox{-0.4\height}{\includegraphics[width=\linewidth]{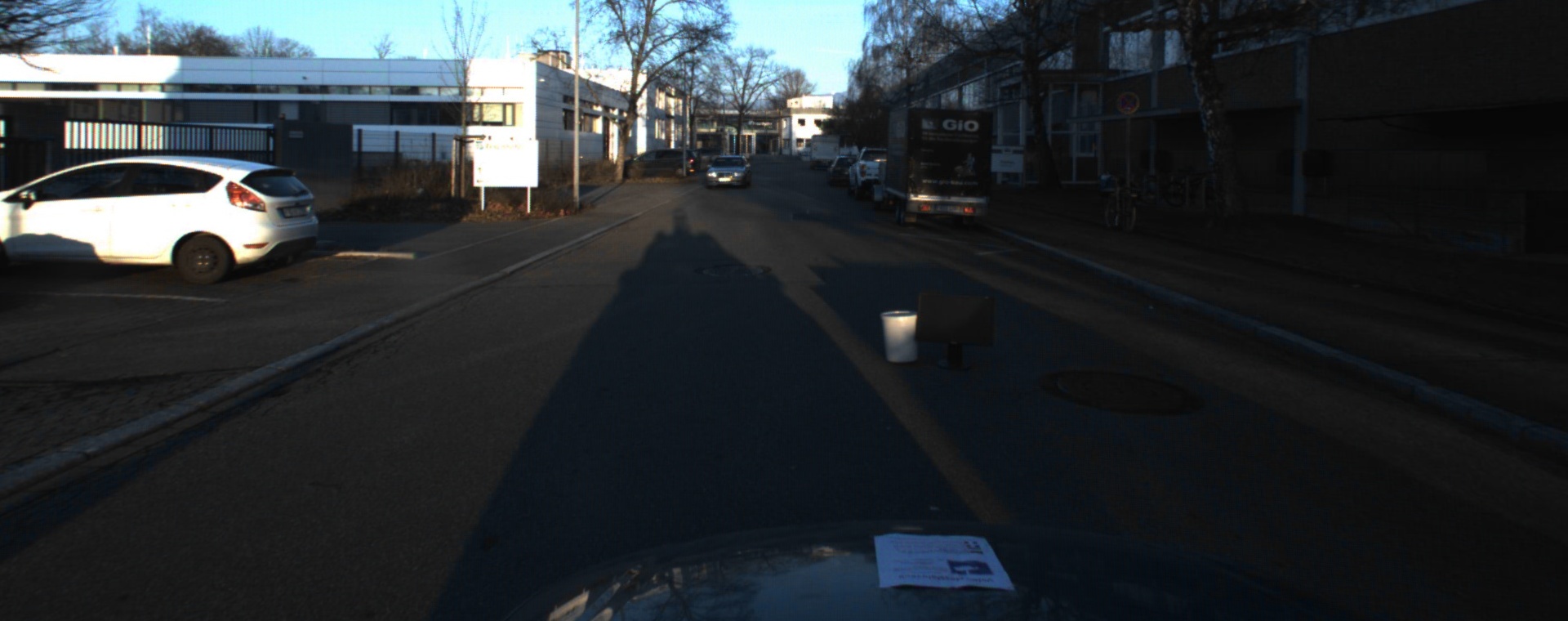}} & \raisebox{-0.4\height}{\includegraphics[width=\linewidth]{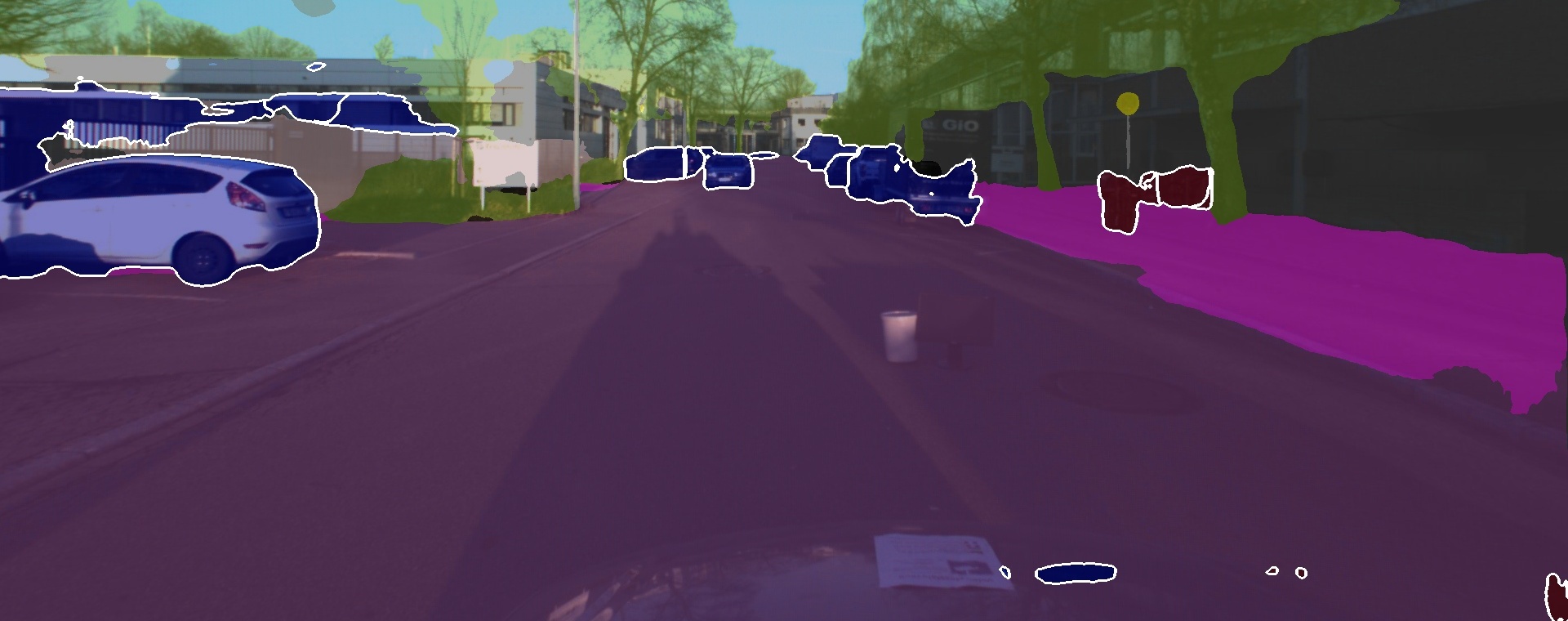}} & \raisebox{-0.4\height}{\includegraphics[width=\linewidth,frame]{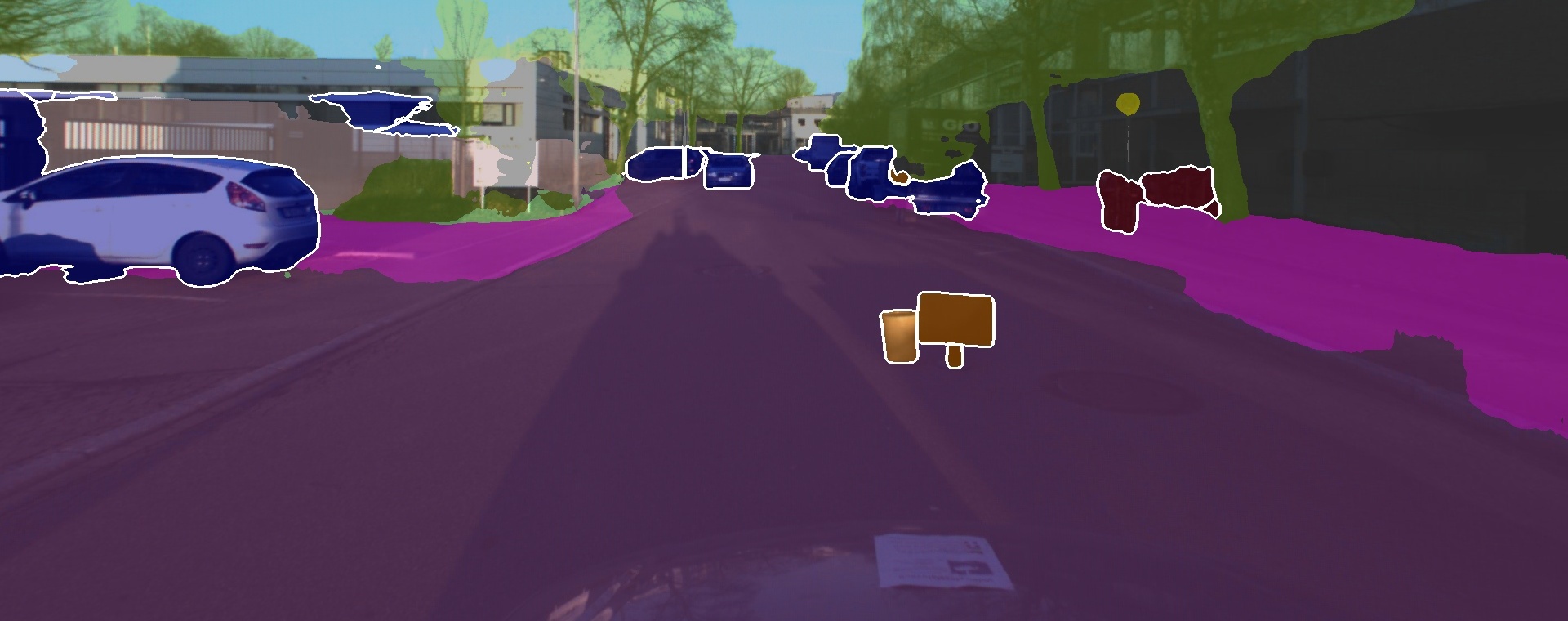}} 
\\

\end{tabular}}
\caption{Qualitative comparison of the PAPS panoptic segmentation network with our PoDS panoptic out-of-distribution segmentation network on real-world scenes featuring out-of-distribution objects. The results highlight the feasibility and importance of panoptic out-of-distribution segmentation for safety-critical applications. } 
\label{fig:svisual_ablation}
\end{figure*}

\subsection{Experimental Setup}
We use (Proposal-free Amodal Panoptic Segmentation) PAPS and PoDS trained on Cityscapes to represent the results while using a panoptic segmentation network and a panoptic out-of-distribution segmentation network, respectively. We train both models using the Cityscapes dataset with images with a resolution of $2048\times1024$. We evaluate on our collected data that has images with a resolution of $1920\times800$. Given the camera resolution difference, as well as the domain difference between the dataset and our collected data obtaining an accurate output on these images is challenging. The low illumination conditions add further complexity to the panoptic out-of-distribution prediction. As a result, our collected data with real-world scenes is suitable to test the quality of the segmentation after deployment.

\subsection{Qualitative Results}

We present qualitative comparisons of PAPS and our proposed PoDS architecture in the supplementary video. Additionally, we present results in \figref{fig:svisual_ablation}. We observe that both models face challenges inherent to panoptic segmentation as well as challenges due to domain, dataset, and camera setup changes. The PAPS model is also unable to identify and segment out-of-distribution objects effectively.
The absence of a mechanism for categorizing novel objects as unknown, results in PAPS misclassifying a suitcase as a car and traffic sign in \figref{fig:svisual_ablation}~(a), and all out-of-distribution (OOD) objects, either part of road or sidewalk in \figref{fig:svisual_ablation}~(b, c, d). In contrast, our PoDS network generalizes from learning specific objects in the training data to unseen objects in the real world. Our network is able to identify more uncommon objects and correctly classify and segment the pixels corresponding to classes such as suitcase, trash bin, and chair. The qualitative results are promising and demonstrate the feasibility of our task as well as the benefits of learning to segment out-of-distribution objects in the scene for safety-critical applications.







\end{document}